\documentclass[letterpaper]{article} % DO NOT CHANGE THIS
\usepackage{aaai25}  % DO NOT CHANGE THIS
\usepackage{times}  % DO NOT CHANGE THIS
\usepackage{helvet}  % DO NOT CHANGE THIS
\usepackage{courier}  % DO NOT CHANGE THIS
\usepackage[hyphens]{url}  % DO NOT CHANGE THIS
\usepackage{graphicx} % DO NOT CHANGE THIS
\usepackage{natbib}  % DO NOT CHANGE THIS AND DO NOT ADD ANY OPTIONS TO IT
\usepackage{caption} % DO NOT CHANGE THIS AND DO NOT ADD ANY OPTIONS TO IT
\usepackage{algorithm}
\usepackage{algorithmic}

\usepackage{newfloat}
\usepackage{listings}
\DeclareCaptionStyle{ruled}{labelfont=normalfont,labelsep=colon,strut=off} % DO NOT CHANGE THIS
\floatstyle{ruled}
\newfloat{listing}{tb}{lst}{}
\floatname{listing}{Listing}
\title{Queryable Prototype Multiple Instance Learning with Vision-Language Models for Incremental Whole Slide Image Classification}
\author{
    Jiaxiang Gou,
    Luping Ji\thanks{Corresponding Author.},
    Pei Liu,
    Mao Ye
}
\affiliations{
    School of Computer Science and Engineering, University of Electronic Science and Technology of China, China\\
    goujiaxiang@std.uestc.edu.cn, jiluping@uestc.edu.cn, yuukilp@std.uestc.edu.cn, maoye@uestc.edu.cn
}
\usepackage{bibentry}
\usepackage{amsmath}
\usepackage{amssymb}
\usepackage{multirow}
\usepackage{booktabs}
\usepackage{makecell}
\usepackage{colortbl}
\usepackage[table]{xcolor}

\newcommand{\rowbg}{yellow!10}

\begin{document}

\maketitle

\begin{abstract}
Whole Slide Image (WSI) classification has very significant applications in clinical pathology, \emph{e.g.}, tumor identification and cancer diagnosis. Currently, most research attention is focused on Multiple Instance Learning (MIL) using \emph{static} WSI datasets. One of the most obvious weaknesses of these methods is that they cannot efficiently preserve and utilize previously learned knowledge. With any new data arriving, classification models are required to be re-trained on both previous and current new data. To overcome this shortcoming and break through traditional vision modality, this paper proposes the first \emph{Vision-Language}-based framework with \emph{Queryable Prototype Multiple Instance Learning} (QPMIL-VL) specially designed for incremental WSI classification. This framework mainly consists of two information processing branches: one is for generating bag-level features by prototype-guided aggregation of instance features, while the other is for enhancing class features through a combination of class ensemble, tunable vector and class similarity loss. The experiments on four public WSI datasets demonstrate that our QPMIL-VL framework is effective for incremental WSI classification and often significantly outperforms other compared methods, achieving state-of-the-art (SOTA) performance. Our source code is publicly available at \textcolor[RGB]{0,67,147}{\emph{https://github.com/can-can-ya/QPMIL-VL}}.
\end{abstract}

% Uncomment the following to link to your code, datasets, an extended version or similar.

% \begin{links}
%     \link{Code}{https://aaai.org/example/code}
%     \link{Datasets}{https://aaai.org/example/datasets}
%     \link{Extended version}{https://aaai.org/example/extended-version}
% \end{links}

\section{Introduction}

Histopathology Whole Slide Image (WSI) is crucial for the diagnosis and treatment of tumor diseases~\cite{song2023artificial}. To model gigapixel WSIs (\textit{e.g.}, $90,000 \times 90,000$ pixels) for training clinical-grade models, many methods based on Multiple Instance Learning (MIL) have been extensively studied in the field of computational pathology (CPATH)~\cite{campanella2019clinical, lu2021data, lin2023interventional, liu10385148}. A shared characteristic of these methods is that they are specially designed for capturing a \textit{static} data distribution (\textit{e.g.}, a given WSI dataset with two lung cancer subtypes), still following conventional machine learning paradigm. However, in the real-world, the distribution of WSI data could be \textit{dynamic} due to the arrival of new datasets or the discovery of emerging cancer types \cite{van2021deep, derakhshani2022lifelonger}. As a result, a well-trained WSI classification model that is suitable for previous data often cannot adapt to new data. Therefore, it must be re-trained on the entire dataset (including both previous and new data), leading to a significant increase in training costs. This presents the necessities of new methods well-tailored to dynamic data distributions.

\begin{figure}[t]
\centering
\includegraphics[width=1\linewidth]{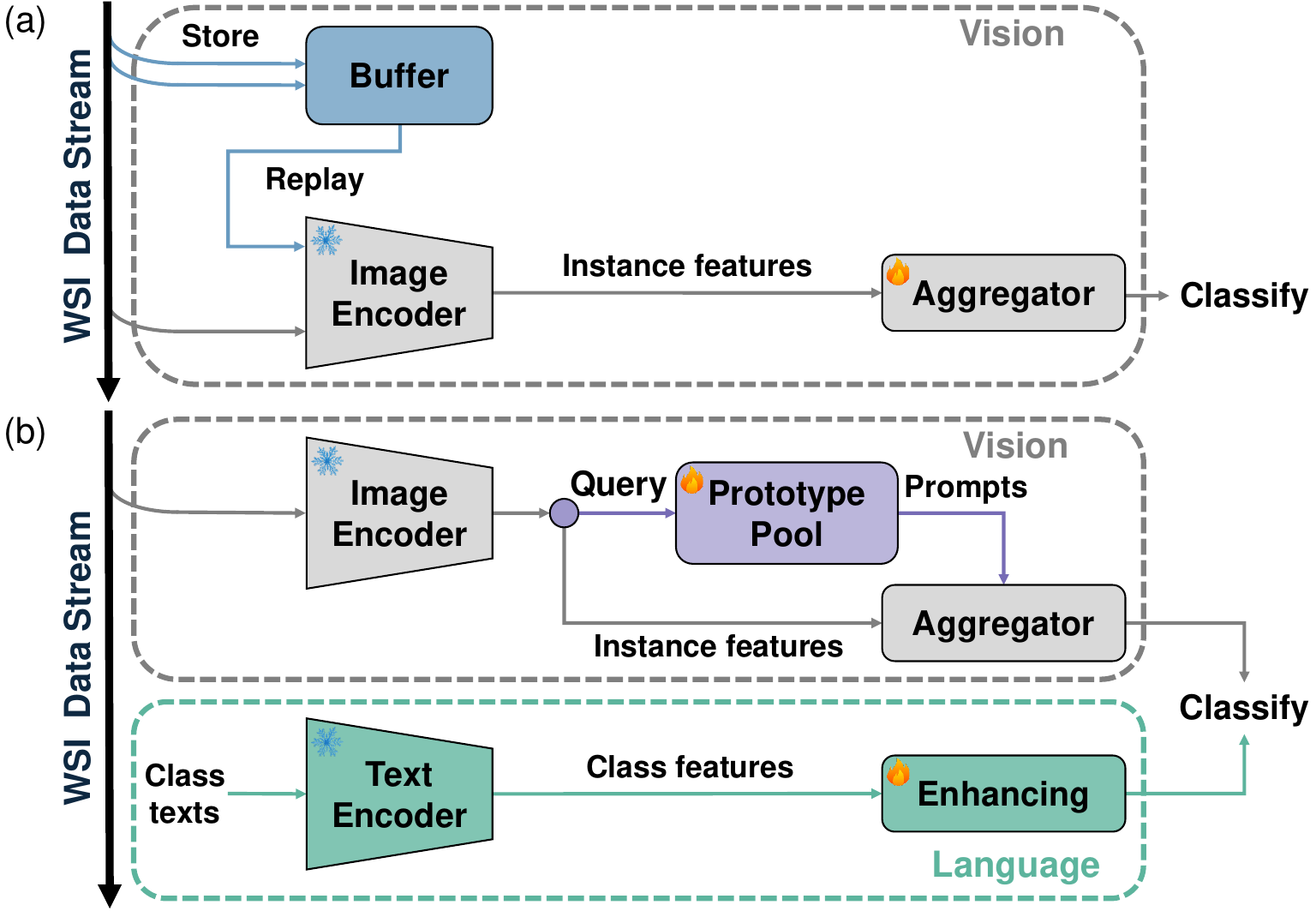}
\caption{(a) Existing visual modality framework for incremental WSI classification with buffer dependency; (b) our proposed Vision-Language-based framework with a queryable prototype pool and class feature enhancement.
}
\label{fig:framwork_intro}
\end{figure}

Incremental Learning (IL)~\cite{li2017learning}, also referred to as continual learning or lifelong learning, is exactly such approach to filling the aforementioned gap. As a new learning paradigm, it can encourage models to maintain the memory stability on old data distributions while adapting to a new one~\cite{wang2024comprehensive}, thus mitigating the notorious \textit{catastrophic forgetting} problem~\cite{mccloskey1989catastrophic}. Owing to such practical benefit, this new paradigm has attracted great attention and has shown strong potential for application in the real-world \cite{li2019learn,caccia2021new,wang2022s}. Similarly, in CPATH, IL strategies are also noticed recently and show promising performance in learning from dynamic WSI data. Concretely, ConSlide~\cite{huang2023conslide} is proposed for the first time to study incremental WSI classification, as illustrated in Fig.~\ref{fig:framwork_intro} (a). \textbf{(1)} Like most existing methods, ConSlide captures image-level features for WSI classification, relying solely on a traditional \textit{vision modality}. \textbf{(2)} Moreover, to retain certain past data, an additional buffer is utilized to enable the model to review previous knowledge while learning new information. This may result in privacy concerns~\cite{shokri2015privacy} and \textit{higher computational costs}, leaving a gap to the resource efficiency goal pursued in IL~\cite{wang2024comprehensive}.

Recent works, \textit{e.g.}, PLIP~\cite{huang2023visual}, Prov-GigaPath~\cite{xu2024whole} and CONCH~\cite{lu2024visual}, have deeply demonstrated the significant potential of Vision-Language Models (VLMs) in CPATH. Breaking through traditional vision modality, this kind of new models could effectively exploit pathology features from both \textit{vision} and \textit{language} modalities. All of these motivate us to leverage foundational pathology VLMs to study incremental learning for WSI classification.

To extend the pure vision frameworks, we propose a new \textbf{Vision-Language-based framework with Queryable Prototype Multiple Instance Learning} (\textbf{QPMIL-VL})  for incremental WSI classification, as shown in Fig.~\ref{fig:framwork_intro} (b). Specifically, \textbf{(1)} in the \textbf{vision} branch, inspired by L2P~\cite{wang2022learning}, we design a Queryable Prototype MIL (QPMIL) module comprising a prototype pool and prototype-guided aggregation. This module encourages the model to incrementally learn a set of prototypes corresponding to each dataset through a prototype-query mechanism, effectively mitigating catastrophic forgetting without relying on extra buffer. \textbf{(2)} In the \textbf{language} branch, we propose a Class Feature Enhancement (CFE) module. We employ CFE to increase the diversity of text descriptions via a class ensemble approach and further refine it with tunable vector and a class similarity loss. Our experiments show that QPMIL-VL could often surpass other state-of-the-art (SOTA) methods by a large margin in incremental WSI classification tasks. The contributions of this work are as follows:

I) To our knowledge, beyond traditional pure vision frameworks, we propose the first Vision-Language-based framework for incremental WSI classification.

II) We devise a new Queryable Prototype Multiple Instance Learning (QPMIL) strategy to alleviate catastrophic forgetting. In this strategy, instance features are matched with a set of prototypes through querying to guide the generation of WSI bag-level feature.

III) Extensive experiments are performed on four public WSI datasets. The results demonstrate the superiority of our QPMIL-VL over the existing methods.

\section{Related Work}

\subsection{Multiple Instance Learning for WSI Classification}

Multiple Instance Learning (MIL) focuses on learning from weakly-annotated data, where only an unknown subset of instances within each input bag is relevant to the label. Due to the unique data characteristics of WSI, the MIL paradigm is widely applied in WSI classification~\cite{campanella2019clinical, li2021dual, shao2021transmil, zhang2022dtfd, zheng2023kernel, liu2024advmil}. Among these methods, ABMIL~\cite{ilse2018attention} proposes an attention-based instance feature aggregation. By leveraging ResNet for instance-level feature extraction, CLAM~\cite{lu2021data} introduces an interpretable weakly-supervised learning method, focused on data-efficient WSI processing. However, most of these studies focus on \textit{static} MIL classification tasks and rarely consider the incremental WSI classifcation with MIL.

\subsection{Incremental Learning}

Recently, Incremental Learning (IL) has gained significant attention, aiming to enable deep models to continually acquire knowledge like humans. IL algorithms can be classified into three main categories. Regularization-based methods~\cite{kirkpatrick2017overcoming, li2017learning} aim to reduce catastrophic forgetting by limiting changes to key parameters but often fall short of optimal results. Architecture-based methods~\cite{rusu2016progressive, mallya2018packnet, li2019learn, ke2020continual} train an independent module for each task but typically apply only to task-incremental learning scenario requiring task identity during inference. Additionally, although prevalent Rehearsal-based methods~\cite{chaudhry2018efficient, chaudhry2019tiny, prabhu2020gdumb, buzzega2020dark, caccia2021new} could achieve SOTA performance on various benchmarks~\cite{parisi2019continual, mai2022online}, these methods rely on additional buffer to store past data. When buffer size is limited, the performance of such these methods deteriorates~\cite{Cha_2021_ICCV}. They will even become inapplicable when historical data is unavailable~\cite{shokri2015privacy}. Currently, most of these studies focus more on \textit{natural images}, not medical gigapixel WSIs, except ConSlide~\cite{huang2023conslide}.

\subsection{Vision-Language Models}

Recent research has made significant success in developing Vision-Language Models (VLMs). For example, CLIP~\cite{radford2021learning} learns SOTA image representations, by the training on 400 million (image, text) pairs. Coca~\cite{yu2022coca} employs contrastive and captioning losses to pre-train a foundation model comprising an image-text encoder-decoder. Additionally, some specialized VLMs like PLIP~\cite{huang2023visual}, Prov-GigaPath~\cite{xu2024whole} and CONCH~\cite{lu2024visual} have been developed in CPATH. These VLMs need large datasets for pre-training, so they could often show excellent generalizability capabilities. Moreover, some VLMs combined with certain fine-tuning methods~\cite{zhou2022learning, yu2023task, gao2024clip} are widely transferred to various downstream tasks~\cite{wang2023attriclip, qu2024rise, li2024generalizable, liu2024interpretable}. At present, to our knowledge, the potential of VLMs for incremental WSI classification remains unexplored.

\section{Preliminaries}

\subsection{Problem Formulation}

Incremental WSI classification requires a model to continuously learn new knowledge from sequential tasks on non-stationary datasets, while retaining the knowledge learned from prior tasks. We define a sequence of WSI datasets $\mathcal{D}=\{\mathcal{D}_{1},\cdots,\mathcal{D}_{T}\}$, in which the $t$-th dataset $\mathcal{D}_t=\{(\boldsymbol{x}_i^t,y_i^t)\}_{i=1}^{n_t}$ contains the tuples of sample $\boldsymbol{x}_i^t$ and its corresponding bag-level label $y_i^t$, where $1\leq t\leq T$ and $n_t$ is the sample number in $\mathcal{D}_t$. When training is made on current dataset $\mathcal{D}_{t_c}$, the data from previous datasets (\textit{i.e.}, $\mathcal{D}_1,\cdots,\mathcal{D}_{t_c-1}$) may be limited or even unavailable.

For a given WSI sample $\boldsymbol{x}_i^t$, in class-incremental learning scenario, the model only uses $\boldsymbol{x}_i^t$ to predict its bag-level label $y_i^t$. In contrast, in task-incremental learning scenario, the model combines sample $\boldsymbol{x}_i^t$ with its corresponding task identity $t$ together to predict label $y_i^t$.

\subsection{VLM Pre-training and Inference}

In our work, we choose a SOTA VLM in pathology, \textit{i.e.}, CONCH~\cite{lu2024visual}, as our image and text encoders.

\textbf{Pre-training.} CONCH uses the contrastive learning on diverse histopathology images, biomedical text, and over $1.17$ million image-caption pairs (not including TCGA Datasets). Based on CoCa~\cite{yu2022coca} framework, CONCH combines an image encoder $\boldsymbol{E}_\text{img}$ with $f(\cdot,\theta)$, a text encoder $\boldsymbol{E}_\text{txt}$ with $g(\cdot,\phi)$, and a multi-modal fusion decoder.

\textbf{Zero-shot Patch Inference.} Since CONCH is pre-trained on patch-level WSI image-text pairs, after training is completed, zero-shot inference can be fulfilled on patch-level samples. Specifically, given a patch $\boldsymbol{p}$ and class text token set $\{\boldsymbol{t}_c\}_{c=1}^{C}$, where $C$ denotes total class number. The classification can be conducted based on the cosine similarity between patch feature $f(\boldsymbol{p},\theta)$ and class feature $g(\boldsymbol{t}_c,\phi)$.

\begin{figure*}[t]
\centering
\includegraphics[width=0.8\textwidth]{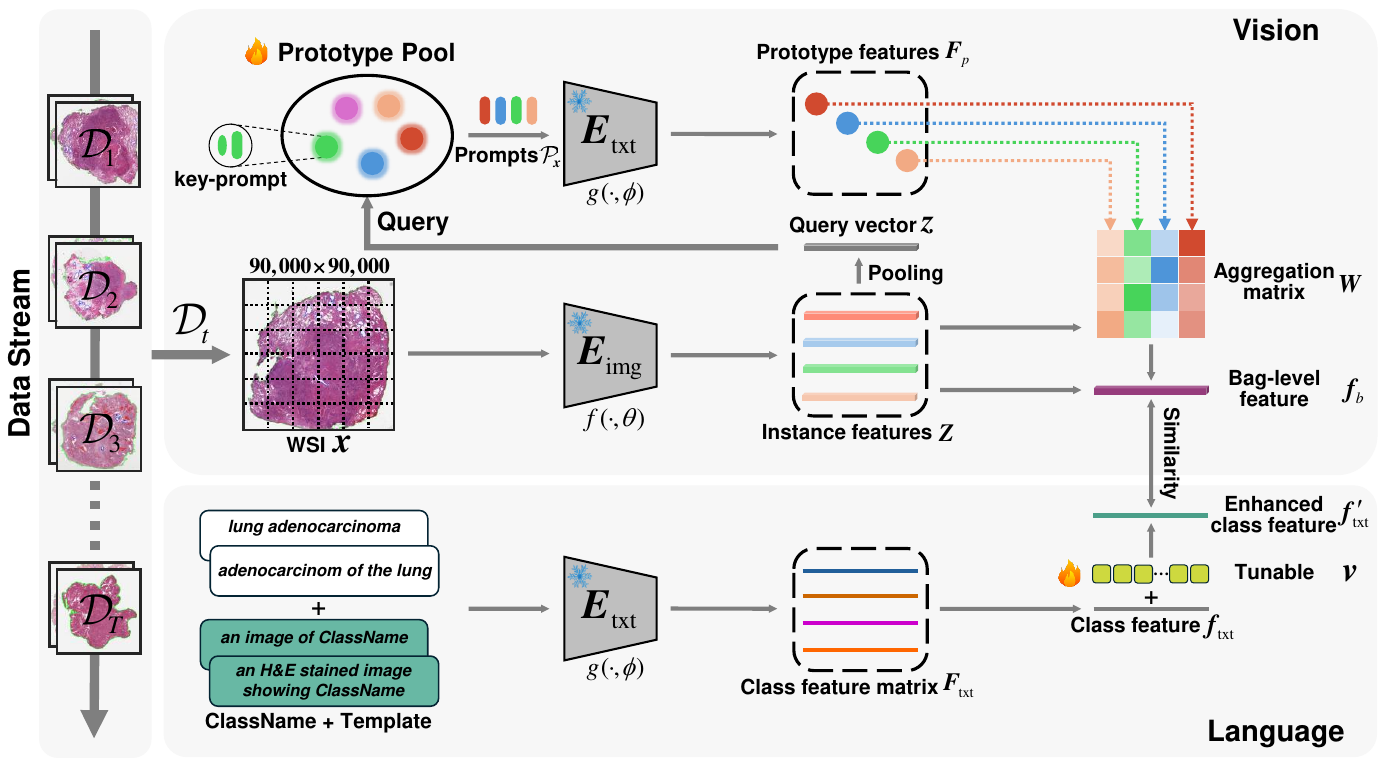}
\caption{The framework of QPMIL-VL. The prompts in the prototype pool enable an efficient incremental learning process by gradually capturing the visual feature descriptions of instance prototypes present in the sequential WSI datasets.
}
\label{fig:framwork}
\end{figure*}

\section{Proposed Method}

To effectively fulfil incremental WSI classification, on traditional MIL framework, we introduce Vision-Language Model and propose a queryable prototype MIL framework, as shown in Fig.~\ref{fig:framwork}. It consists of two branches, one for generating bag-level feature (QPMIL), and the other for enhancing class text feature (CFE). In inference, the features from both branches will be used to compute classification probability.

\subsection{Queryable Prototype Multiple Instance Learning (QPMIL)}

As usual, one of the most important objectives for WSI classification is to generate discriminative bag-level feature representation. However, unlike traditional static classification, our incremental classification task requires that all subsequent model learning specifically avoids causing catastrophic impacts on the tasks learned earlier.

Moreover, it has been proved that the instances in different WSIs, even in different WSI datasets, could often appear high similarities, in terms of cell shape, staining response, and tissue arrangement~\cite{song2024morphological}.Therefore, similar instances, even if they are from different datasets, could be clustered into the same instance prototype. In view of this, to generate more effective WSI bag-level feature, we introduce VLM and instance-level feature prototypes, designing a queryable prototype pool and prototype-guided aggregation.

\textbf{Prototype Pool.} Specifically, the prototype pool consists of $M$ (\textit{key}, \textit{prompt}) prototype pairs $\{(\boldsymbol{k}_i,\boldsymbol{P}_i)\}_{i=1}^M$ initialized using a \textit{Normal Distribution}, $\boldsymbol{k}_i\in\mathbb{R}^{D_f},\boldsymbol{P}_i\in\mathbb{R}^{L_{\boldsymbol{P}}\times D_e}$, where $D_f$ is the output feature dimension of encoder $\boldsymbol{E}_\text{img}$ and $\boldsymbol{E}_\text{txt}$, $D_e$ represents the embedding dimension of $\boldsymbol{E}_\text{txt}$ and $L_{\boldsymbol{P}}$ indicates the total number of learnable vectors in $\boldsymbol{P}_i$. In a prototype pair, the \textit{key} works as a prototype identifier, used by WSIs for querying, while the \textit{prompt} acts as a prototype descriptor, utilized to describe the specific visual feature of instance prototype within WSIs.

Given a WSI sample $\boldsymbol{x}$ from $\mathcal{D}_t$, it is first divided into $n$ non-overlapping patches $\{\boldsymbol{p}_i\}_{i=1}^n$, \textit{i.e.}, instances. Then, all instances are encoded by image encoder $\boldsymbol{E}_\text{img}$ with $f(\cdot,\theta)$, generating $n$ instance features $\boldsymbol{Z}\in\mathbb{R}^{n\times D_f}$, as shown in Fig.~\ref{fig:framwork}. Next, we apply \textit{MaxPooling} to features $\boldsymbol{Z}$ to obtain the query vector of sample $\boldsymbol{x}$, \textit{i.e.}, $\boldsymbol{z}=MaxPooling(\boldsymbol{Z})\in\mathbb{R}^{D_f}$. \textit{MeanPooling} is also tried and its experimental result is provided in our Supplementary Material.

Subsequently, a query metric function $q:\mathbb{R}^{D_f}\times\mathbb{R}^{D_f}\to\mathbb{R}$ is defined to evaluate how close the query vector $\boldsymbol{z}$ matches the \textit{key} of each prototype pair. We employs the cosine distance for this evaluation. Finally, we are able to find the most matched prototype \textit{key}s through
\begin{equation}
    \mathcal{K}_{\boldsymbol{x}}=Top\text{-}N^{min}\{q(\boldsymbol{z},\boldsymbol{k}_i)\}_{i=1}^M,
    \label{eq:choose_top_n_v1}
\end{equation}
where $Top\text{-}N^{min}$ denotes the operation of choosing the top-$N$ prototype \textit{key}s with minimal $\boldsymbol{z}$-\textit{key} distances. However, in prototype matching on different WSI datasets during incremental training by Eq.~\ref{eq:choose_top_n_v1}, we observe that almost all samples tend to choose the same set of $N$ prototypes from pool. This could usually lead to catastrophic forgetting in incremental learning. To effectively avoid this tendency, we define a penalty factor $p_i^{t_c}$ to optimize prototype choosing:
\begin{equation}
\left\{\begin{aligned}
    \mathcal{K}_{\boldsymbol{x}}&=Top\text{-}N^{min}\{q(\boldsymbol{z},\boldsymbol{k}_i)\cdot p_i^{t_c}\}_{i=1}^M\\
    P_{t_c}&=\frac{1}{t_c-1}\sum_{t=1}^{t_c-1}F_t,t_c\geqslant2,
\label{eq:choose_top_n_v2}
\end{aligned}\right.
\end{equation}
where $t_c$ is the index of current WSI dataset $\mathcal{D}_{t_c}$, $F_t=[f_1^t,\cdots,f_M^t]$ denotes the matching frequency table of $M$ prototype \textit{key}s on dataset $\mathcal{D}_t$ and $p_i^{t_c}$ represents the $i$-th element of penalty table $P_{t_c}$. In essence, this penalty factor amplifies the distances of $\boldsymbol{z}$ to all the prototype \textit{key}s that have high frequency matching on those datasets prior to $\mathcal{D}_{t_c}$.

\textbf{Prototype-guided Aggregation.} After \textit{key} matching, the \textit{prompt}s $\mathcal{P}_{\boldsymbol{x}}=\{\boldsymbol{P}_i^{\boldsymbol{x}}\}_{i=1}^N$ of top-$N$ prototype pairs will be further fed into text encoder $\boldsymbol{E}_\text{txt}$ to generate $N$ prototype features $\boldsymbol{F}_p\in\mathbb{R}^{N\times D_f}$.

Next, we compute the cosine similarity matrix $\boldsymbol{S}\in\mathbb{R}^{n\times N}$ between instance features $\boldsymbol{Z}$ and prototype features $\boldsymbol{F}_p$. Since each column in $\boldsymbol{S}$ exactly corresponds to a prototype, a \textit{Softmax} normalization along column direction is performed to obtain a weighted aggregation matrix $\boldsymbol{W}\in\mathbb{R}^{n\times N}$. Finally, we obtain the prototype-guided bag-level features through $\boldsymbol{W}$, and their average yields the bag-level feature $\boldsymbol{f}_b\in\mathbb{R}^{D_f}$ for the WSI sample $\boldsymbol{x}$. The procedure can be represented as
\begin{equation}
    \boldsymbol{f}_b=mean(Softmax(\boldsymbol{Z}\times {\boldsymbol{F}_p}^{\top})^{\top}\times \boldsymbol{Z}).
    \label{eq:cal_bag_fea}
\end{equation}

Ideally, in incremental training, different datasets will match a non-overlapped set of prototype pairs. In practice, some prototype pairs could often be shared with low frequencies by different datasets. Our method doesn't impose a strict independence of matched prototype pairs. Instead, it allows WSI datasets to choose their optimal prototypes, providing more flexibility and adaptability.

\subsection{Class Feature Enhancement (CFE)}

As shown in Fig.~\ref{fig:framwork}, class text feature enhancing branch is designed to generate an enhanced class feature $\boldsymbol{f'}_\text{txt}$.

\textbf{Class Ensemble.} For a specific subtype of cancer, such as ``lung adenocarcinoma'', it can also be expressed as ``adenocarcinoma of the lung''. Therefore, we adopt class ensemble to obtain better class feature. For each cancer subtype, we combine several common class names and templates to obtain $m$ different class text descriptions. All text descriptions are encoded by text encoder $\boldsymbol{E}_\text{txt}$ with $g(\cdot,\phi)$ to obtain initial class feature matrix $\boldsymbol{F}_\text{txt}\in\mathbb{R}^{m\times D_f}$, and by averaging $\boldsymbol{F}_\text{txt}$, we get the class feature $\boldsymbol{f}_\text{txt}\in\mathbb{R}^{D_f}$ for this subtype. The entire calculation process can be expressed as
\begin{equation}
    \boldsymbol{f}_\text{txt}=mean(g(\boldsymbol{T},\phi)),
    \label{eq:class_ensemble}
\end{equation}
where $\boldsymbol{T}\in\mathbb{R}^{m\times L}$ represents the text tokens corresponding to the text descriptions and $L$ is the length of text token.

\textbf{Tunable Vector.} To further refine class feature $\boldsymbol{f}_\text{txt}$, referring to the TaskRes~\cite{yu2023task}, we design a tunable (\textit{i.e.}, learnable) vector $\boldsymbol{v}\in\mathbb{R}^{D_f}$ of the same dimension for each class. The final enhanced class feature $\boldsymbol{f'}_\text{txt}\in\mathbb{R}^{D_f}$ is calculated through
\begin{equation}
    \boldsymbol{f'}_\text{txt}=\boldsymbol{f}_\text{txt}+\alpha\cdot\boldsymbol{v},
    \label{eq:tunable_v}
\end{equation}
where $\alpha$ (default constant $0.5$) is an additional amplitude factor to control the superposition scale of vector $\boldsymbol{v}$ on feature $\boldsymbol{f}_\text{txt}$.

\subsection{Prediction and Optimization}

\textbf{Class Prediction.} The probability of predicting the WSI sample $\boldsymbol{x}$ as the class $y$ can be computed as
\begin{equation}
    p(y|\boldsymbol{x})=
    \frac{e^{\tau\cdot\langle \boldsymbol{f}_b,[\boldsymbol{f'}_\text{txt}]_y\rangle}}
    {\sum_{c=1}^C e^{\tau\cdot\langle \boldsymbol{f}_b,[\boldsymbol{f'}_\text{txt}]_c\rangle}},
    \label{eq:wsi_classification}
\end{equation}
where $C$ indicates the total number of classes, $[\boldsymbol{f'}_\text{txt}]_c$ represents the $c$-th class feature, $\tau$ is a temperature parameter learned by CONCH~\cite{lu2024visual} and $\langle\cdot,\cdot\rangle$ denotes the cosine similarity.

\textbf{Optimization Objective.} To appropriately reduce the similarity between features of different classes, we firstly define a \textit{class similarity loss}, which is calculated through
\begin{equation}
    \mathcal{L}_S=\frac{2}{C\cdot(C-1)}\sum_{i=1}^C\sum_{j=i+1}^C\langle[\boldsymbol{f'}_\text{txt}]_i,[\boldsymbol{f'}_\text{txt}]_j\rangle+1.
    \label{eq:loss_s}
\end{equation}

Moreover, there are another two losses. One is \textit{classification loss}. Based on Eq.~\ref{eq:wsi_classification}, it is computed by
\begin{equation}
    \mathcal{L}_C=-\text{log}
    \frac{e^{\tau\cdot\langle \boldsymbol{f}_b,[\boldsymbol{f'}_\text{txt}]_y\rangle}}
    {\sum_{c=1}^C e^{\tau\cdot\langle \boldsymbol{f}_b,[\boldsymbol{f'}_\text{txt}]_c\rangle}}.
    \label{eq:loss_c}
\end{equation}

\begin{table*}[t]
\centering
\resizebox{1\textwidth}{!}{
\begin{tabular}{c|l|lccc|c|l}
    \toprule
    
    IL Type & Method & \makecell[c]{ACC ($\uparrow$)} & Upper-bound Ratio ($\uparrow$) & Forgetting ($\downarrow$) & BWT ($\uparrow$) & Params & \makecell[c]{Masked ACC ($\uparrow$)} \\
    
    \midrule
    
    \multirow{2}*{Baseline} 
    & JointTrain (Upper)  & 0.908 $\pm$ 0.022 & 1.000 $\pm$ 0.000 & - & - & \multirow{2}*{0.502} & 0.937 $\pm$ 0.022 \\
    
    & FineTune (Lower) & 0.308 $\pm$ 0.045 $^{***}$ & 0.336 $\pm$ 0.045 & 0.841 $\pm$ 0.054 & -0.841 $\pm$ 0.054 & & 0.844 $\pm$ 0.038 $^{***}$ \\
    
    \midrule
    
    \multirow{2}*{Regularization-based} & EWC~\cite{kirkpatrick2017overcoming} & 0.309 $\pm$ 0.051 $^{***}$ & 0.338 $\pm$ 0.055 & 0.836 $\pm$ 0.064 & -0.836 $\pm$ 0.064 & \multirow{2}*{0.502} & 0.840 $\pm$ 0.035 $^{***}$ \\
    
    & LwF~\cite{li2017learning} & 0.378 $\pm$ 0.081 $^{***}$ & 0.412 $\pm$ 0.083 & 0.731 $\pm$ 0.116 & -0.731 $\pm$ 0.116 & & 0.916 $\pm$ 0.031 \\
    
    \midrule
    
    \multirow{6}*{Rehearsal-based} & A-GEM/30~\cite{chaudhry2018efficient} & 0.436 $\pm$ 0.058 $^{***}$ & 0.477 $\pm$ 0.070 & 0.670 $\pm$ 0.072 & -0.670 $\pm$ 0.072 & \multirow{4}*{0.502} & 0.879 $\pm$ 0.045 $^{**}$ \\
    
    & ER-ACE/30~\cite{caccia2021new} & 0.666 $\pm$ 0.049 $^{***}$ & 0.732 $\pm$ 0.051 & 0.075 $\pm$ 0.057 & -0.070 $\pm$ 0.059 & & 0.917 $\pm$ 0.023 \\
    
    & DER++/30~\cite{buzzega2020dark} & 0.749 $\pm$ 0.055 $^{***}$ & 0.823 $\pm$ 0.059 & 0.219 $\pm$ 0.059 & -0.214 $\pm$ 0.062 & & 0.904 $\pm$ 0.029 $^{**}$ \\
    
    & ER/30~\cite{chaudhry2019tiny} & 0.790 $\pm$ 0.040 $^{***}$ & 0.870 $\pm$ 0.049 & 0.186 $\pm$ 0.044 & -0.186 $\pm$ 0.044 & & 0.894 $\pm$ 0.036 $^{**}$ \\
    
    \cline{2-8}
    
    & ConSlide/30~\cite{huang2023conslide} & \textcolor{gray!95}{0.659 $\pm$ 0.022} & - & \textcolor{gray!95}{0.076 $\pm$ 0.030} & \textcolor{gray!95}{-0.075 $\pm$ 0.030} & 39.446 & \textcolor{gray!95}{0.861 $\pm$ 0.017} \\
    
    & ER/100~\cite{chaudhry2019tiny} & 0.827 $\pm$ 0.028 $^{***}$ & 0.910 $\pm$ 0.034 & 0.143 $\pm$ 0.034 & -0.142 $\pm$ 0.032 & 0.502 & 0.918 $\pm$ 0.028 \\
    
    \midrule
    
    \multirow{3}*{Pre-trained VLM-based} 
    & AttriCLIP~\cite{wang2023attriclip} & 0.616 $\pm$ 0.056 $^{***}$ & 0.677 $\pm$ 0.071 & 0.285 $\pm$ 0.059 & -0.285 $\pm$ 0.059 & \textbf{0.111} & 0.844 $\pm$ 0.026 $^{***}$ \\
    
    & MI-Zero~\cite{lu2023visual} & 0.839 $\pm$ 0.034 $^{***}$ & 0.927 $\pm$ 0.044 & - & - & - & 0.909 $\pm$ 0.019 $^{**}$ \\
    
    \rowcolor{\rowbg} \cellcolor{white} & \textbf{QPMIL-VL (ours)} & \textbf{0.890 $\pm$ 0.021} & \textbf{0.982 $\pm$ 0.032} & \textbf{0.027 $\pm$ 0.014} & \textbf{-0.027 $\pm$ 0.014} & 0.365 & \textbf{0.930 $\pm$ 0.018} \\
    
    \bottomrule
\end{tabular}
}
\caption{Main results in class-incremental learning scenario on forward-order training. The best performances are highlighted as \textbf{bold}. ``Params'' represents the number of learnable parameters (M). Gray numbers are directly cited from published paper. ``/30'' and ``/100'' mean buffer $30$ WSIs and buffer $100$ WSIs, respectively. We present the Masked ACC reflecting task-incremental learning scenario for reference. $^{*}$/$^{**}$/$^{***}$ denote there are significant differences (paired t-test \textit{p}-value $<$ $0.05$/$0.01$/$0.001$) between the best performance and others.
}
\label{tab:il_result_forward}
\end{table*}

The other is \textit{matching loss} to make the matched \textit{key}s $\mathcal{K}_{\boldsymbol{x}}=\{\boldsymbol{k}_i^{\boldsymbol{x}}\}_{i=1}^N$ closer to the query vector $\boldsymbol{z}$ of WSI sample $\boldsymbol{x}$. It is defined as 
\begin{equation}
    \mathcal{L}_M=\frac{1}{N}\sum_{i=1}^Nq(\boldsymbol{z},\boldsymbol{k}_i^{\boldsymbol{x}}).
    \label{eq:loss_m}
\end{equation}

To optimize our QPMIL-VL model in training, based on the three losses above, \textit{total loss} $\mathcal{L}_T$ is computed by
\begin{equation}
    \mathcal{L}_T=\mathcal{L}_C+\lambda\cdot\mathcal{L}_M+\beta\cdot\mathcal{L}_S,
    \label{eq:loss_o}
\end{equation}
where balance factors $\lambda$ and $\beta$ are set to constant $0.5$. Therefore, our optimization objective is to minimize $\mathcal{L}_T$.

\section{Experiments}

Following the setup of previous representative work~\cite{van2019three}, we mainly evaluate QPMIL-VL in the class-incremental learning scenario and additionally report the results in the task-incremental learning scenario.

\subsection{Experimental Details}

\textbf{Datasets.} Following ConSlide~\cite{huang2023conslide}, we use four public WSI datasets from The Cancer Genome Atla (TCGA) repository: non-small cell lung carcinoma (NSCLC), invasive breast carcinoma (BRCA), renal cell carcinoma (RCC) and esophageal carcinoma (ESCA). With the arrival of each dataset, we add two additional classes to the model for training and evaluation.

We use CLAM~\cite{lu2021data} to crop non-overlapping $256\times256$ patches from the segmented tissue at $10\times$ magnification. Then, pre-trained image encoder $\boldsymbol{E}_\text{img}$ in  CONCH~\cite{lu2024visual} is used to extract instance features.

\textbf{Evaluation Metrics.} We assess all incremental learning methods with Average Accuracy (ACC), Upper-bound Ratio, Forgetting and Backward Transfer (BWT)~\cite{lopez2017gradient, hayes2018new, fini2022self}. Besides, we provide the Masked Average Accuracy (Masked ACC) metric, which is used for reference only to evaluate performance in task-incremental learning scenario by masking the logits of irrelevant classes. All results are reported by the mean and standard deviation of ten-fold cross-validation.

\textbf{Training Details.} In all experiments, we set the number of training epoch to $12$ for each dataset and use the Adam~\cite{kingma2014adam} optimizer with $\text{lr}=0.001$ and $\text{weight\_decay}=0.0005$. The mini-batch sizes on forward-order and reverse-order training are set to $16$ and $8$, respectively. For QPMIL-VL, we set $M=20,N=5$ and $L_{\boldsymbol{P}}=24$. All experiments are conducted on a machine with two NVIDIA GeForce RTX $3090$ GPUs.

Refer to our Supplementary Material for more details.

\subsection{Comparison Results}

\textbf{Compared Methods.} It is necessary to compare with the upper and lower bounds of IL methods. We employ the classic ABMIL~\cite{ilse2018attention} as the aggregator for the vision branch, serving as a \textbf{Baseline}. Its performances by jointly training on all datasets (JointTrain, \textit{non-incremental}) are taken as upper bound, and those by naively fine-tuning sequential datasets (FineTune) are taken as lower bound. In addition, we compare \textbf{Regularization-based} methods: EWC~\cite{kirkpatrick2017overcoming} and LwF~\cite{li2017learning}; \textbf{Rehearsal-based} methods: A-GEM~\cite{chaudhry2018efficient}, ER~\cite{chaudhry2019tiny}, DER++~\cite{buzzega2020dark}, ER-ACE~\cite{caccia2021new} and ConSlide~\cite{huang2023conslide}; and \textbf{Pre-trained VLM-based} methods: AttriCLIP~\cite{wang2023attriclip} and MI-Zero~\cite{lu2023visual}. To ensure a fair comparison, all methods are adapted to VLM. Specially, they use the same instance features and text encoder $\boldsymbol{E}_\text{txt}$ in CONCH.

\begin{table*}[t]
\centering
\resizebox{1\textwidth}{!}{
\begin{tabular}{c|l|lccc|l}
    \toprule
    
    IL Type & Method & \makecell[c]{ACC ($\uparrow$)} & Upper-bound Ratio ($\uparrow$) & Forgetting ($\downarrow$) & BWT ($\uparrow$) & \makecell[c]{Masked ACC ($\uparrow$)} \\
    
    \midrule
    
    \multirow{2}*{Baseline} & JointTrain (Upper) & 0.908 $\pm$ 0.022 & 1.000 $\pm$ 0.000 & - & - & 0.937 $\pm$ 0.022 \\
    
    & FineTune (Lower) & 0.234 $\pm$ 0.008 $^{***}$ & 0.262 $\pm$ 0.010 & 0.927 $\pm$ 0.023 & -0.927 $\pm$ 0.023 & 0.803 $\pm$ 0.060 $^{***}$ \\
    
    \midrule
    
    \multirow{2}*{Regularization-based} 
    & EWC~\cite{kirkpatrick2017overcoming} & 0.235 $\pm$ 0.011 $^{***}$ & 0.264 $\pm$ 0.011 & 0.928 $\pm$ 0.020 & -0.928 $\pm$ 0.020 & 0.833 $\pm$ 0.069 $^{***}$ \\
    
    & LwF~\cite{li2017learning} & 0.236 $\pm$ 0.016 $^{***}$ & 0.265 $\pm$ 0.012 & 0.908 $\pm$ 0.030 & -0.908 $\pm$ 0.030 & 0.900 $\pm$ 0.041 \\
    
    \midrule
    
    \multirow{5}*{Rehearsal-based} 
    & A-GEM/30~\cite{chaudhry2018efficient} & 0.536 $\pm$ 0.047 $^{***}$ & 0.591 $\pm$ 0.053 & 0.527 $\pm$ 0.067 & -0.527 $\pm$ 0.067 & 0.872 $\pm$ 0.026 $^{***}$ \\
    
    & ER-ACE/30~\cite{caccia2021new} & 0.703 $\pm$ 0.049 $^{***}$ & 0.777 $\pm$ 0.053 & 0.281 $\pm$ 0.062 & -0.279 $\pm$ 0.063 & 0.889 $\pm$ 0.041 $^{*}$ \\
    
    & DER++/30~\cite{buzzega2020dark} & 0.684 $\pm$ 0.055 $^{***}$ & 0.755 $\pm$ 0.063 & 0.310 $\pm$ 0.069 & -0.307 $\pm$ 0.072 & 0.910 $\pm$ 0.043 \\
    
    & ER/30~\cite{chaudhry2019tiny} & 0.644 $\pm$ 0.028 $^{***}$ & 0.711 $\pm$ 0.033 & 0.387 $\pm$ 0.050 & -0.386 $\pm$ 0.049 & 0.901 $\pm$ 0.035 \\
    
    %\cline{2-7}
    
    & ConSlide/30~\cite{huang2023conslide} & \textcolor{gray!95}{0.499 $\pm$ 0.025} & - & \textcolor{gray!95}{0.058 $\pm$ 0.032} & \textcolor{gray!95}{-0.021 $\pm$ 0.039} & \textcolor{gray!95}{0.854 $\pm$ 0.039} \\
    
    \midrule
    
    \multirow{3}*{Pre-trained VLM-based} 
    & AttriCLIP~\cite{wang2023attriclip} & 0.694 $\pm$ 0.058 $^{***}$ & 0.766 $\pm$ 0.061 & 0.207 $\pm$ 0.063 & -0.207 $\pm$ 0.063 & 0.861 $\pm$ 0.019 $^{***}$ \\
    
    & MI-Zero~\cite{lu2023visual} & 0.839 $\pm$ 0.034 & 0.927 $\pm$ 0.044 & - & - & 0.909 $\pm$ 0.019 \\
    
    \rowcolor{\rowbg} \cellcolor{white} & \textbf{QPMIL-VL (ours)} & \textbf{0.859 $\pm$ 0.032} & \textbf{0.946 $\pm$ 0.028} & \textbf{0.064 $\pm$ 0.031} & \textbf{-0.064 $\pm$ 0.031} & \textbf{0.925 $\pm$ 0.018} \\
    
    \bottomrule
\end{tabular}
}
\caption{Main results in class-incremental learning scenario on reverse-order training.
}
\label{tab:il_result_reverse}
\end{table*}

\textbf{Results on Forward-order Training.} This set of comparisons on NSCLC$\rightarrow$BRCA$\rightarrow$RCC$\rightarrow$ESCA is shown in Tab.~\ref{tab:il_result_forward}. From this table, we have three empirical findings.
 
\textbf{(1) Our QPMIL-VL method is consistently superior to all other methods, with parameters only slightly larger than the smallest one.} QPMIL-VL achieves an ACC of $0.890$ $\pm$ $0.021$ with an Upper-bound Ratio as high as $0.982$ $\pm$ $0.032$, winning SOTA performance metrics. Its Upper-bound Ratio is even $5.5\%$ higher than that of MI-Zero, the second-best method. Additionally, our model has a small learnable parameter size of $0.365$M, which is only slightly larger than the smallest size of $0.111$M in AttriCLIP. However, in terms of performance metrics, our method is far superior to the lowest-parameter method, \textit{i.e.}, AttriCLIP. Besides, in task-incremental experiments, our Masked ACC could reach up to $0.930$ $\pm$ $0.018$, which is also significantly higher than any achieved by other compared methods.
 
\textbf{(2) Two regularization-based methods are almost ineffective in WSI class-incremental learning scenario.} For example, as a regularization-based method, LwF only achieves an ACC of $0.378$ $\pm$ $0.081$, which is slightly higher than the lower bound of $0.308$ $\pm$ $0.045$.
 
\textbf{(3) Most rehearsal-based and pre-trained VLM-based methods could achieve good classification results.} The rehearsal-based DER++ acquires an ACC of $0.749$ $\pm$ $0.055$ with an Upper-bound Ratio of $0.823$ $\pm$ $0.059$, and the pre-trained VLM-based MI-Zero obtains an ACC of $0.839$ $\pm$ $0.034$ with an Upper-bound Ratio of $0.927$ $\pm$ $0.044$. Their Upper-bound Ratio results exceed $80\%$. Therefore, these two types of methods are believed to be more effective than the former, consistent with natural image observations. Notably, although AttriCLIP performs well in the natural image-based IL setting, its performance in this task is lower than some rehearsal-based methods, reflecting the unique challenges of analyzing gigapixel WSI. We consider the main reasons are the poor bag-level feature (obtained through \textit{MaxPooling}) and the overly similar class features derived from the same set of prompts.

\begin{table}[t]
\centering
\resizebox{0.8\columnwidth}{!}{
\begin{tabular}{c|ccc|cc}
    \toprule
    
     \multirow{2}*{QPMIL} & \multicolumn{3}{c|}{CFE} & \multirow{2}*{ACC ($\uparrow$)} & \multirow{2}*{Forgetting ($\downarrow$)} \\
    
    \cline{2-4}
      
    & TV & CE & CSL & & \\
    
    \midrule
    
    & & & & 0.308 $\pm$ 0.045 & 0.841 $\pm$ 0.054 \\
    
    $\checkmark$ & & & & 0.720 $\pm$ 0.029 & 0.158 $\pm$ 0.025 \\
    
    $\checkmark$ & $\checkmark$ & & & 0.747 $\pm$ 0.039 & 0.149 $\pm$ 0.037\\
    
    $\checkmark$ & $\checkmark$ & $\checkmark$ & & 0.874 $\pm$ 0.033 & 0.052 $\pm$ 0.030 \\
    
    \rowcolor{\rowbg} $\checkmark$ & $\checkmark$ & $\checkmark$ & $\checkmark$ & \textbf{0.890 $\pm$ 0.021} & \textbf{0.027 $\pm$ 0.014} \\

    \bottomrule
\end{tabular}
}
\caption{Ablation studies on main components (TV: Tunable Vector, CE: Class Ensemble, CSL: Class Similarity Loss).
}
\label{tab:il_ablation_1}
\end{table}

\begin{table}[t]
\centering
% \small
\resizebox{0.8\columnwidth}{!}{
\begin{tabular}{l|cc}
    \toprule
    
    \textbf{Ablation} & ACC ($\uparrow$) & Forgetting ($\downarrow$) \\
    
    \midrule
    
    w/o key & 0.675 $\pm$ 0.029 & 0.332 $\pm$ 0.034 \\
    
    w/o matching penalty & 0.659 $\pm$ 0.029 & 0.356 $\pm$ 0.048 \\
    
    \midrule
    
    \rowcolor{\rowbg} \textbf{QPMIL-VL} & \textbf{0.890 $\pm$ 0.021} & \textbf{0.027 $\pm$ 0.014} \\

    \bottomrule
\end{tabular}
}
\caption{Ablation studies on prototype pool.
}
\label{tab:il_ablation_2}
\end{table}

\textbf{Results on Reverse-order Training.} In incremental classification, the training order of the dataset could influence the model's performance. The results on reverse-order training (ESCA$\rightarrow$RCC$\rightarrow$BRCA$\rightarrow$NSCLC) are shown in Tab.~\ref{tab:il_result_reverse}. By comparison, we have three observations, which are almost the same as those from Tab.~\ref{tab:il_result_forward}.

\textbf{(1)} Our QPMIL-VL method is consistently superior to all other methods in both class-incremental and task-incremental learning scenarios, maintaining a SOTA position. \textbf{(2)} Regularization-based methods are still ineffective in class-incremental learning scenario due to suffering from severe catastrophic forgetting. \textbf{(3)} Most rehearsal-based and pre-trained VLM-based methods often achieve good classification performance.

\textbf{Comparison between Forward-order and Reverse-order Training.}~Additionally, we observe that on reverse-order training, the performance of most methods is noticeably lower than on forward-order training. For example, our QPMIL-VL decreases to an ACC of $0.859$ $\pm$ $0.032$. In contrast, its ACC is $0.890$ $\pm$ $0.021$ on forward-order training (as shown in Tab.~\ref{tab:il_result_forward}). One possible explanation to this decline is that in IL, later datasets can affect earlier ones. ESCA, with only $150$ slides, is the first to be learned, while NSCLC, with $965$ slides, is the last. NSCLC is dominant in the total number of training samples, which could cause a greater influence on the earlier datasets, leading to more forgetting.

\subsection{Ablation and Analysis}

\textbf{Ablation Studies.}~The following are several ablation analyses of QPMIL-VL.

\textbf{I) on Main Components}~To analyze the impact of the main components on performance, we conduct a set of ablation studies on forward-order training. The results are shown in Tab.~\ref{tab:il_ablation_1}, where the first row represents our baseline (FineTune). Two obvious findings can be observed as follows.

\textbf{(1)} Each component in QPMIL-VL contributes to the performance improvement. For example, starting from the baseline, QPMIL can significantly boost the ACC from $0.308$ $\pm$ $0.045$ to $0.720$ $\pm$ $0.029$. Similarly, incrementally applying Class Similarity Loss raises the ACC from $0.874$ $\pm$ $0.033$ to $0.890$ $\pm$ $0.021$. \textbf{(2)} Among all components, both QPMIL and Class Ensemble have a more significant impact on performance than the other two. This is because the prototype query mechanism in QPMIL effectively mitigates catastrophic forgetting, while the Class Ensemble significantly optimizes class feature encoding in the VLM by increasing the diversity of class text descriptions. These comparisons imply that each component is effective.

\textbf{II) on Prototype Pool}~Furthermore, we conduct another set of ablation studies specifically on the queryable prototype pool. Our experimental results are shown in Tab.~\ref{tab:il_ablation_2}.

\textbf{(1)} ``\textbf{w/o key}'' indicates the removal of the \textit{key} from prototype pair, implying that incremental training always uses the same set of \textit{prompt}s. In this situation, the knowledge from different datasets is encoded into the same set of \textit{prompt}s, leading to catastrophic forgetting and a significant performance decline (ACC dropping to $0.675$ $\pm$ $0.029$). \textbf{(2)} ``\textbf{w/o matching penalty}'' presents that only  Eq.~\ref{eq:choose_top_n_v1} is used for matching prototype pairs. Under this ablation, the ACC metric decreases to $0.659$ $\pm$ $0.029$ because different datasets tend to match the same set of prototypes during training, which consequently results in catastrophic forgetting. Only when both key and matching penalty are applied together in our QPMIL-VL does the ACC reach its highest value of $0.890$ $\pm$ $0.021$. This ablation demonstrates that both are beneficial for incremental classification.

\textbf{Further Performance Analysis.}~Additionally, we conduct additional experiments to further analyze our method.

\begin{figure}[htbp]
\centering
\includegraphics[width=1\linewidth]{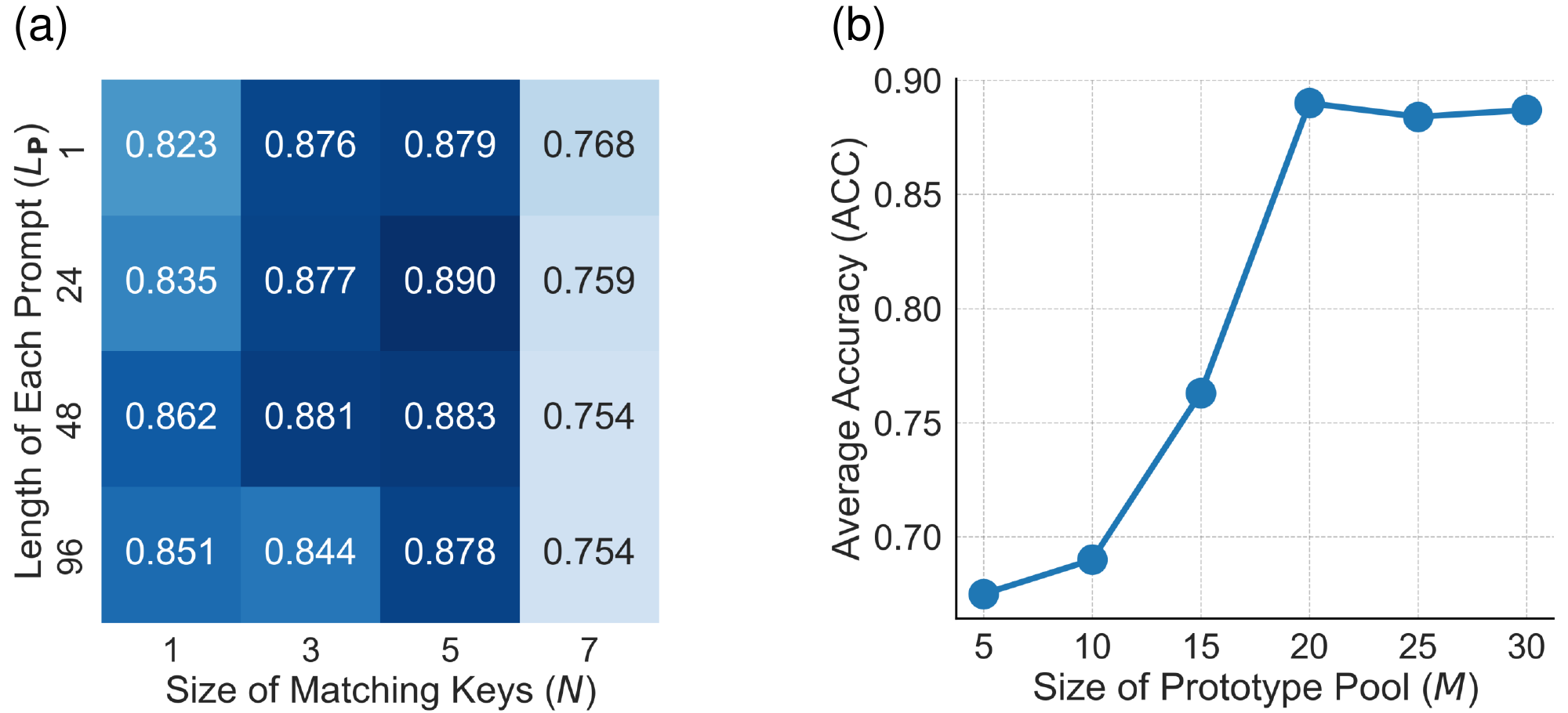}
\caption{(a) ACC w.r.t length of each prompt ($L_{\boldsymbol{P}}$) and size of matching keys ($N$), size of prototype pool $M=20$; (b) ACC w.r.t $M$, $L_{\boldsymbol{P}}=24$ and $N=5$.
}
\label{fig:hyperparam}
\end{figure}

\textbf{I) Effect of Hyper-parameters}~As shown in Fig.~\ref{fig:hyperparam}, there are three important parameters in the prototype pool. The results indicate that inappropriate prototype pool capacity ((a): $N=7$, (b): $M<20$) may result in knowledge interference and too few \textit{prompt}s ((a): $N=1$) make it difficult to describe the various instance prototype visual features present in WSIs.

\begin{figure}[htbp]
\centering
\includegraphics[width=1\linewidth]{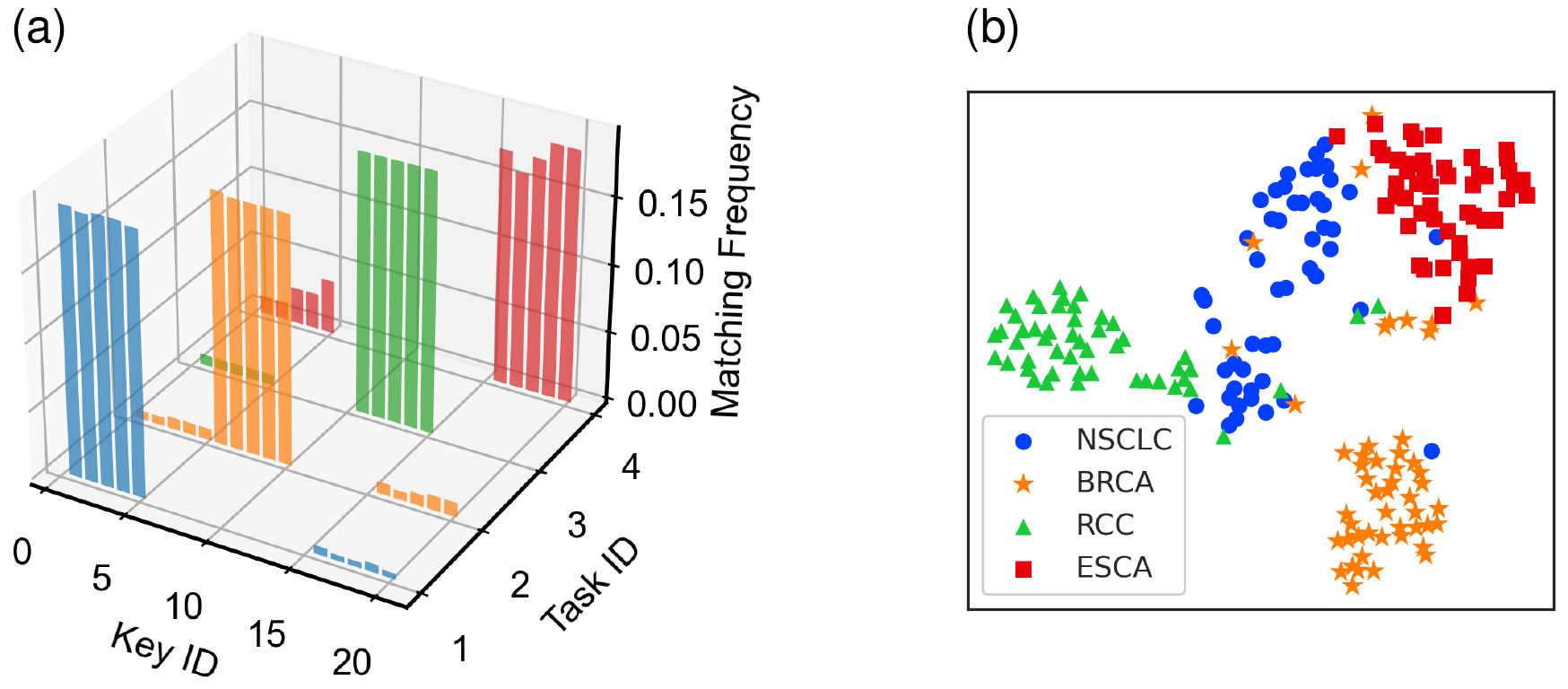}
\caption{(a) Prototype key matching frequency histogram; (b) prototype feature visualization.
}
\label{fig:key_matching}
\end{figure}

\textbf{II) Prototype Key Matching}~To analyze the matching results of \textit{key}s on incremental datasets, we compute the matching frequency of each prototype pair in the first fold of experiments, as illustrated in Fig.~\ref{fig:key_matching} (a). It can be observed that each dataset tends to stably match $5$ \textit{key}s. Moreover, the $5$ \textit{key}s predominantly matched by different datasets are often non-overlapping due to the matching penalty.

\textbf{III) Prototype Feature Visualization}~As depicted in Fig.~\ref{fig:key_matching} (b), we present the distribution visualization of $200$ prototype features using t-SNE~\cite{van2008visualizing}. These features are derived from the ten-fold cross-validation experiments conducted on four datasets, \textit{i.e.}, $200=10\times4\times5$. Their distributions show that most prototypes learned from the same dataset exhibit clear clustering properties. From another perspective, they also imply that our model can learn distinct prototype features from different datasets. Additionally, from Fig.~\ref{fig:key_matching} (a) and (b), we observe a small number of outliers. They can be regarded as shared prototypes.

\begin{figure}[htbp]
\centering
\includegraphics[width=1\linewidth]{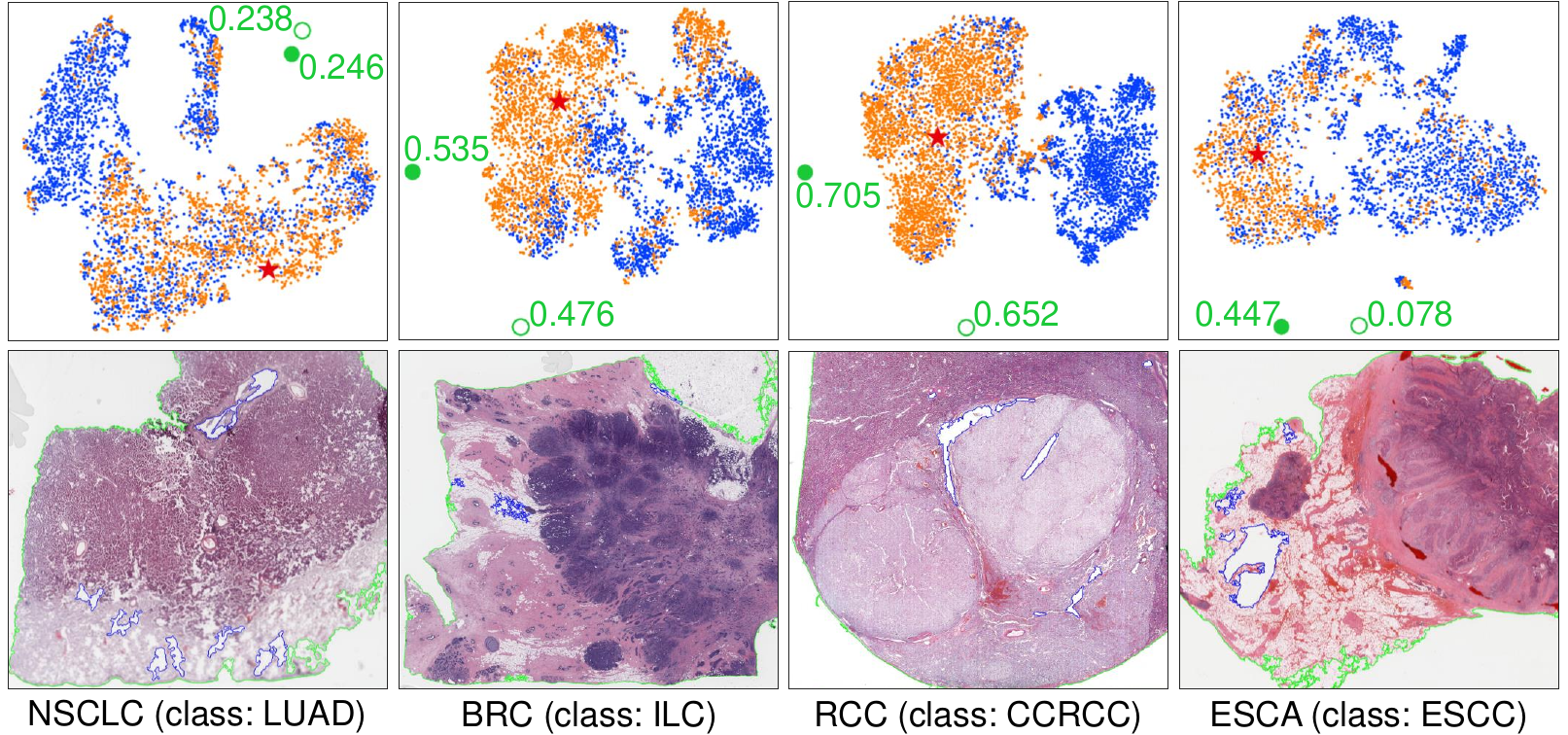}
\caption{Class features visualization, only one class per dataset. Red \textcolor{red}{$\star$} is the center, \textit{i.e.}, the average of features, for the instances of interest (orange points, determined by the cosine similarity between instance features and learned prototype features). Blue points are non-interest instance features. Green \textcolor{green}{$\circ$} and \textcolor{green}{$\bullet$} denote the cosine similarities before and after feature enhancement, respectively.
}
\label{fig:class_feature_enhancement}
\end{figure}

\textbf{IV) Class Feature Enhancement Analysis}~To show the effectiveness of CFE, we visualize four classes of WSI samples and their corresponding class features, as shown in Fig.~\ref{fig:class_feature_enhancement}. More experimental results can be found in our Supplementary Material.

From the alteration in feature similarity, we can observe that the class features are noticeably improved by the enhancement. For example, in the case of the class sample ESCA-ESCC, the initial cosine similarity between the center point feature and the class feature is only $0.078$, but it increases to $0.447$ after enhancement, resulting in an improvement of $0.369$. This alteration suggests that the CFE is effective in enhancing the quality of class features.

\section{Conclusions}

This paper proposes the first Vision-Language-based framework for incremental WSI classification, QPMIL-VL, breaking through traditional vision modality. By a pool of learnable prototype pairs, it aggregates the instance features from pre-trained image encoder to generate bag-level feature under the guidance of matched prototype \textit{prompt}s. Class probability is predicted by the cosine similarity between bag-level feature and enhanced class feature. Extensive experiments demonstrate that our method and its components are effective. It obviously outperforms other methods in incremental WSI classification, achieving SOTA results.

\section{Acknowledgments}

This work is supported by the National Natural Science Foundation of China (NSFC) under Grant No.62476049.

\bibliography{main}

\clearpage
{

\appendix

\section{Supplementary Material}

\subsection{}

\subsection{A. Dataset Details}

The benchmark consists of four public WSI datasets from TCGA repository: non-small cell lung carcinoma (NSCLC), invasive breast carcinoma (BRCA), renal cell carcinoma (RCC) and esophageal carcinoma (ESCA), as shown in Tab.~\ref{tab:il_datasets}.

\begin{table}[h]
\centering
\resizebox{1\columnwidth}{!}{
\begin{tabular}{llcc}
    \toprule
    
    Dataset & Tumor Type & Cases & Slides \\
    
    \midrule
    
    \multirow{2}*{NSCLC}    & Lung adenocarcinoma (LUAD) & 449 & 512 \\
    
                            & Lung squamous cell carcinoma (LUSC) & 419 & 453 \\
                            
    \midrule
    
    \multirow{2}*{BRCA}     & Invasive ductal (IDC) & 713 & 761 \\
    
                            & Invasive lobular carcinoma (ILC) & 178 & 191 \\
                            
    \midrule
    
    \multirow{2}*{RCC}      & Clear cell renal cell carcinoma (CCRCC) & 493 & 499 \\
    
                            & Papillary renal cell carcinoma (PRCC) & 245 & 269 \\
                            
    \midrule
    
    \multirow{2}*{ESCA}     & Esophageal adenocarcinoma (ESAD) & 64 & 64 \\
    
                            & Esophageal squamous cell carcinoma (ESCC) & 84 & 86 \\
                            
    \bottomrule
\end{tabular}
}
\caption{The statistics of incremental WSI analysis benchmark.
}
\label{tab:il_datasets}
\end{table}

\subsection{B. Metric Details}

After completing the incremental training, we can obtain an Accuracy performance matrix, as illustrated in Tab.~\ref{tab:performance_matrix}. Subsequently, the relevant metrics reported in our experiments are calculated using this performance matrix.

\begin{table}[h]
\centering
\resizebox{1\columnwidth}{!}{
\begin{tabular}{c|cccc}
    \toprule
    
    \multirow{2}*{\makecell[c]{After \\ Training}} & \multicolumn{4}{c}{Test on} \\
     
    \cline{2-5}
    
    & Dataset $1$ & Dataset $2$ & $\cdots$ & Dataset $T$ \\
                            
    \midrule
    
    Dataset $1$ & $R_{1,1}$ & - & $\cdots$ & - \\
    
    Dataset $2$ & $R_{2,1}$ & $R_{2,2}$ & $\cdots$ & - \\
    
    $\cdots$ & $\cdots$ & $\cdots$ & $\cdots$ & $\cdots$ \\
    
    Dataset $T$ & $R_{T,1}$ & $R_{T,2}$ & $\cdots$ & $R_{T,T}$ \\
    
    \midrule
    
    JointTrain (Upper) & $R_{joint,1}$ & $R_{joint,2}$ & $\cdots$ & $R_{joint,T}$ \\
                            
    \bottomrule
\end{tabular}
}
\caption{The Accuracy performance matrix during incremental training.
}
\label{tab:performance_matrix}
\end{table}

\textbf{ACC.} Average Accuracy (ACC) represents the model's average performance across all datasets, which is calculated by
\begin{equation}
    ACC=\frac{1}{T}\sum_{t=1}^TR_{T,t}.
    \label{eq:acc}
\end{equation}

\textbf{Upper-bound Ratio.} Upper-bound Ratio enables an incrementally trained model to be compared relative to a jointly trained model (upper bound). It can be computed by
\begin{equation}
    Upper\text{-}bound\ Ratio=\frac{1}{T}\sum_{t=1}^T\frac{R_{T,t}}{R_{joint,t}}.
    \label{eq:ratio}
\end{equation}

\textbf{Forgetting.} Forgetting quantifies how much knowledge the model has forgotten about previous datasets, which is defined as
\begin{equation}
    Forgetting=\frac{1}{T-1}\sum_{t=1}^{T-1}\max_{i\in\{1,\cdots,T\}}R_{i,t}-R_{T,t}.
    \label{eq:forgetting}
\end{equation}

\textbf{BWT.} Backward Transfer (BWT) reflects the model's ability to mitigate catastrophic forgetting (\textit{i.e.}, memory stability). The calculation formula is as follows:
\begin{equation}
    BWT=\frac{1}{T-1}\sum_{t=1}^{T-1}R_{T,t}-R_{t,t}.
    \label{eq:bwt}
\end{equation}

\subsection{C. More Details of Class Ensemble}

Tab.~\ref{tab:class_ensemble} provides all the class names and templates used in our implementation of the class ensemble. They are adapted from those used in CONCH.

\begin{table*}[t]
\centering
\resizebox{1\textwidth}{!}{
\begin{tabular}{l|c|cl}
    \toprule
    
    Template & Dataset & Tumor Type & ClassName \\
     
    \midrule
    
    \multirow{30}*{\makecell[l]{``ClassName'' \\ ``a photomicrograph showing ClassName'' \\ ``a photomicrograph of ClassName'' \\ ``an image of ClassName'' \\ ``an image showing ClassName'' \\ ``an example of ClassName'' \\ ``ClassName is shown'' \\ ``this is ClassName'' \\ ``there is ClassName'' \\ ``a histopathological image showing ClassName'' \\ ``a histopathological image of ClassName'' \\ ``a histopathological photograph of ClassName'' \\ ``a histopathological photograph showing ClassName'' \\ ``shows ClassName'' \\ ``presence of ClassName'' \\ ``ClassName is present'' \\ ``an H\&E stained image of ClassName'' \\ ``an H\&E stained image showing ClassName'' \\ ``an H\&E image showing ClassName'' \\ ``an H\&E image of ClassName'' \\ ``ClassName, H\&E stain'' \\ ``ClassName, H\&E''}} & \multirow{6}*{NSCLC} & \multirow{3}*{LUAD} & ``lung adenocarcinoma'' \\
    
    & & & ``adenocarcinoma of the lung'' \\
    
    & & & ``LUAD'' \\
    
    \cline{3-4}
    
    & & \multirow{3}*{LUSC} & ``lung squamous cell carcinoma'' \\
    
    & & & ``squamous cell carcinoma of the lung'' \\
    
    & & & ``LUSC'' \\
    
    \cline{2-4}
    
    & \multirow{10}*{BRCA} & \multirow{5}*{IDC} & ``invasive ductal carcinoma'' \\
    
    & & & ``breast invasive ductal carcinoma'' \\
    
    & & & ``invasive ductal carcinoma of the breast'' \\
    
    & & & ``invasive carcinoma of the breast, ductal pattern'' \\
    
    & & & ``breast IDC'' \\
    
    \cline{3-4}
    
    & & \multirow{5}*{ILC} & ``invasive lobular carcinoma'' \\
    
    & & & ``breast invasive lobular carcinoma'' \\
    
    & & & ``invasive lobular carcinoma of the breast'' \\
    
    & & & ``invasive carcinoma of the breast, lobular pattern'' \\
    
    & & & ``breast ILC'' \\
    
    \cline{2-4}
    
    & \multirow{8}*{RCC} & \multirow{4}*{CCRCC} & ``clear cell renal cell carcinoma'' \\
    
    & & & ``renal cell carcinoma, clear cell type'' \\
    
    & & & ``renal cell carcinoma of the clear cell type'' \\
    
    & & & ``clear cell RCC'' \\
    
    \cline{3-4}
    
    & & \multirow{4}*{PRCC} & ``papillary renal cell carcinoma'' \\
    
    & & & ``renal cell carcinoma, papillary type'' \\
    
    & & & ``renal cell carcinoma of the papillary type'' \\
    
    & & & ``papillary RCC'' \\
    
    \cline{2-4}
    
    & \multirow{6}*{ESCA} & \multirow{3}*{ESAD} & ``esophageal adenocarcinoma'' \\
    
    & & & ``adenocarcinoma of the esophageal'' \\
    
    & & & ``ESAD'' \\
    
    \cline{3-4}
    
    & & \multirow{3}*{ESCC} & ``esophageal squamous cell carcinoma'' \\
    
    & & & ``squamous cell carcinoma of the esophageal'' \\
    
    & & & ``ESCC'' \\
                            
    \bottomrule
\end{tabular}
}
\caption{The class names and templates used in our class ensemble.
}
\label{tab:class_ensemble}
\end{table*}

\subsection{D. Analysis on Accuracy Performance Matrix}

The performance variations of deep models across different datasets during incremental learning cannot be thoroughly analyzed solely by relying on final metrics such as ACC. Therefore, we present detailed evaluation results of QPMIL-VL on both forward-order and reverse-order incremental training in Tables~\ref{tab:performance_matrix_forward} and~\ref{tab:performance_matrix_reverse}, respectively. Through our analysis, there are three findings.

\begin{table}[htbp]
\centering
\resizebox{1\columnwidth}{!}{
\begin{tabular}{c|cccc}
    \toprule
    
    \multirow{2}*{\makecell[c]{After \\ Training}} & \multicolumn{4}{c}{Test on} \\
     
    \cline{2-5}
    
    & NSCLC & BRCA & RCC & ESCA \\
                            
    \midrule
    
    NSCLC & 0.906 $\pm$ 0.027 & - & - & - \\
    
    BRCA & 0.895 $\pm$ 0.027 & 0.878 $\pm$ 0.024 & - & - \\
    
    RCC & 0.892 $\pm$ 0.024 & 0.876 $\pm$ 0.025 & 0.931 $\pm$ 0.022 & - \\
    
    ESCA & 0.837 $\pm$ 0.046 & 0.868 $\pm$ 0.029 & 0.929 $\pm$ 0.023 & 0.928 $\pm$ 0.083 \\
    
    \midrule
    
    JointTrain & 0.891 $\pm$ 0.041 & 0.908 $\pm$ 0.029 & 0.921 $\pm$ 0.045 & 0.913 $\pm$ 0.055 \\
                            
    \bottomrule
\end{tabular}
}
\caption{The Accuracy performance matrix of QPMIL-VL on forward-order training.
}
\label{tab:performance_matrix_forward}
\end{table}

\begin{table}[htbp]
\centering
\resizebox{1\columnwidth}{!}{
\begin{tabular}{c|cccc}
    \toprule
    
    \multirow{2}*{\makecell[c]{After \\ Training}} & \multicolumn{4}{c}{Test on} \\
     
    \cline{2-5}
    
    & ESCA & RCC & BRCA & NSCLC \\
                            
    \midrule
    
    ESCA & 0.967 $\pm$ 0.045 & - & - & - \\
    
    RCC & 0.967 $\pm$ 0.045 & 0.932 $\pm$ 0.029 & - & - \\
    
    BRCA & 0.955 $\pm$ 0.060 & 0.930 $\pm$ 0.030 & 0.873 $\pm$ 0.030 & - \\
    
    NSCLC & 0.807 $\pm$ 0.092 & 0.913 $\pm$ 0.039 & 0.860 $\pm$ 0.028 & 0.856 $\pm$ 0.029 \\
    
    \midrule
    
    JointTrain & 0.913 $\pm$ 0.055 & 0.921 $\pm$ 0.045 & 0.908 $\pm$ 0.029 & 0.891 $\pm$ 0.041 \\
                            
    \bottomrule
\end{tabular}
}
\caption{The Accuracy performance matrix of QPMIL-VL on reverse-order training.
}
\label{tab:performance_matrix_reverse}
\end{table}

\textbf{(1) Forgetting phenomena are commonly observed in the old datasets.} On forward-order training, the Accuracy of NSCLC decreases from $0.906$ $\pm$ $0.027$ to $0.837$ $\pm$ $0.046$, while on reverse-order training, the Accuracy of ESCA drops from $0.967$ $\pm$ $0.045$ to $0.807$ $\pm$ $0.092$. This reveals the prevalent phenomenon of forgetting during the incremental learning process.

\textbf{(2) Our QPMIL-VL demonstrates competitiveness in static WSI classification when compared to the baseline (JointTrain).} For example, on forward-order training, after QPMIL-VL is trained on the fourth dataset, ESCA, it reaches an Accuracy of $0.928$ $\pm$ $0.083$, which is higher than the $0.913$ $\pm$ $0.055$ of JointTrain. However, due to forgetting caused by incremental learning, QPMIL-VL shows a noticeable drop in performance on earlier datasets, so its overall performance ends up being lower than that of JointTrain.

\textbf{(3) The primary reason for the performance degradation on reverse-order training compared to forward-order training lies in the greater impact of learning the fourth dataset on the performance of the first dataset.} After training on the fourth dataset, the Accuracy of the first dataset decreases by $0.055$ ($0.892$ $\pm$ $0.024$ $\to$ $0.837$ $\pm$ $0.046$) and $0.148$ ($0.955$ $\pm$ $0.060$ $\to$ $0.807$ $\pm$ $0.092$) on forward-order and reverse-order training, respectively. During reverse training, NSCLS significantly affects the performance of ESCA, this may be because the NSCLS sample size ($965$) is about six times larger than that of ESCA ($150$), making it easier for NSCLS to interfere with ESCA. In contrast, during forward training, ESCA has a much smaller impact on NSCLS.

\subsection{E. Query Vector Generation Strategies}

\textbf{MaxPooling vs MeanPooling.}~To investigate the impact of different pooling operations on the performance of QPMIL-VL when generating query vector, we conduct comparative experiments using \textit{MaxPooling} and \textit{MeanPooling}. Tab.~\ref{tab:pool} presents the experimental results.

The results show that QPMIL-VL performs better overall with \textit{MaxPooling} compared to \textit{MeanPooling}, particularly on forward-order training (the ACC increases from $0.877$ $\pm$ $0.034$ to $0.890$ $\pm$ $0.021$). These results reflect that the query vectors obtained through \textit{MaxPooling} could possess greater discriminative ability across different datasets (refer to Fig.~\ref{fig:pool}).

\begin{table*}[t]
\centering
\resizebox{1\textwidth}{!}{
\begin{tabular}{c|l|cccc|c}
    \toprule
    
    Training Order & Pooling Type & ACC ($\uparrow$) & Upper-bound Ratio ($\uparrow$) & Forgetting ($\downarrow$) & BWT ($\uparrow$) & Masked ACC ($\uparrow$) \\
    
    \midrule
    
     & MeanPooling & 0.877 $\pm$ 0.034 & 0.966 $\pm$ 0.039 & 0.045 $\pm$ 0.019 & -0.044 $\pm$ 0.019 & 0.929 $\pm$ 0.027 \\
    
    \rowcolor{\rowbg} \cellcolor{white} \multirow{-2}*{Forward-order} & MaxPooling & \textbf{0.890 $\pm$ 0.021} & \textbf{0.982 $\pm$ 0.032} & \textbf{0.027 $\pm$ 0.014} & \textbf{-0.027 $\pm$ 0.014} & \textbf{0.930 $\pm$ 0.018} \\
    
    \midrule

    & MeanPooling & 0.859 $\pm$ 0.042 & 0.946 $\pm$ 0.042 & 0.070 $\pm$ 0.054 & -0.070 $\pm$ 0.055 & \textbf{0.935 $\pm$ 0.015} \\
    
    \rowcolor{\rowbg} \cellcolor{white} \multirow{-2}*{Reverse-order}  & MaxPooling & \textbf{0.859 $\pm$ 0.032} & \textbf{0.946 $\pm$ 0.028} & \textbf{0.064 $\pm$ 0.031} & \textbf{-0.064 $\pm$ 0.031} & 0.925 $\pm$ 0.018 \\
    
    \bottomrule
\end{tabular}
}
\caption{The comparative results of \textit{MaxPooling} and \textit{MeanPooling} on both forward-order and reverse-order training.
}
\label{tab:pool}
\end{table*}

\subsection{F. Prototype Visualization on Whole Slide Images}

The pre-defined \textit{prompt}s in the prototype pool are designed to describe the visual features of the instance prototypes of interest present in WSIs. To validate this design principle, we calculate the cosine similarity between the prompt features (\textit{i.e.}, prototype features) learned by QPMIL-VL and the instance features, and use this similarity as the basis to determine the prototype category to which each instance belongs. The experimental results are presented in Fig.~\ref{fig:nsclc},~\ref{fig:brca},~\ref{fig:rcc} and~\ref{fig:esca}.

\begin{figure}[t]
\centering
\includegraphics[width=1\linewidth]{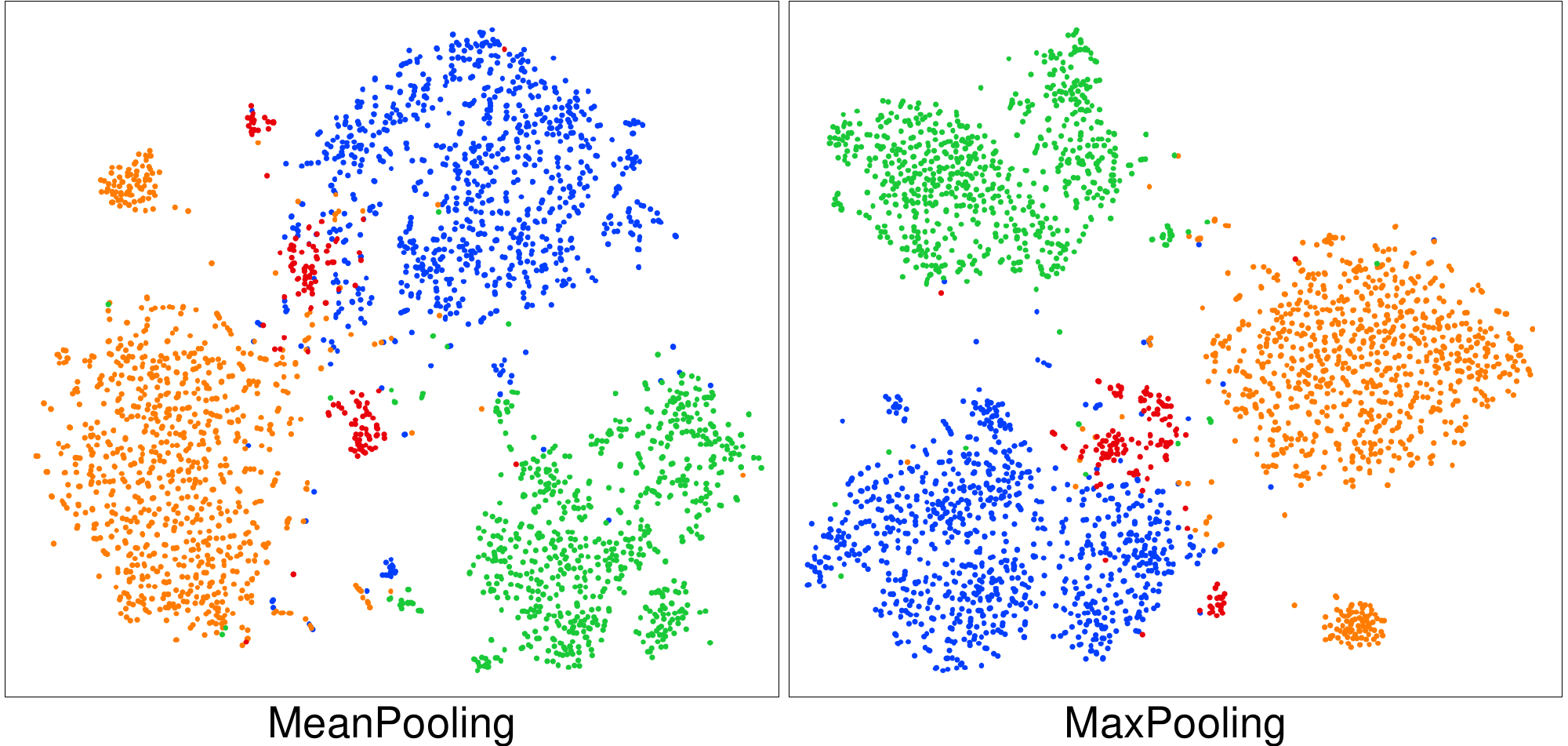}
\caption{The visualization of the query vectors from the four datasets obtained through \textit{MeanPooling} (left) and \textit{MaxPooling} (right).
}
\label{fig:pool}
\end{figure}

By observing the results, we can clearly see that \textit{prompt}s can correctly identify the target instance prototypes (\textit{i.e.}, the lesion areas in the WSIs). Additionally, different \textit{prompt}s can locate different instance prototypes to some extent. This experimental phenomenon is consistent with our expectations.

\begin{figure*}[t]
\centering
\includegraphics[width=0.99\textwidth]{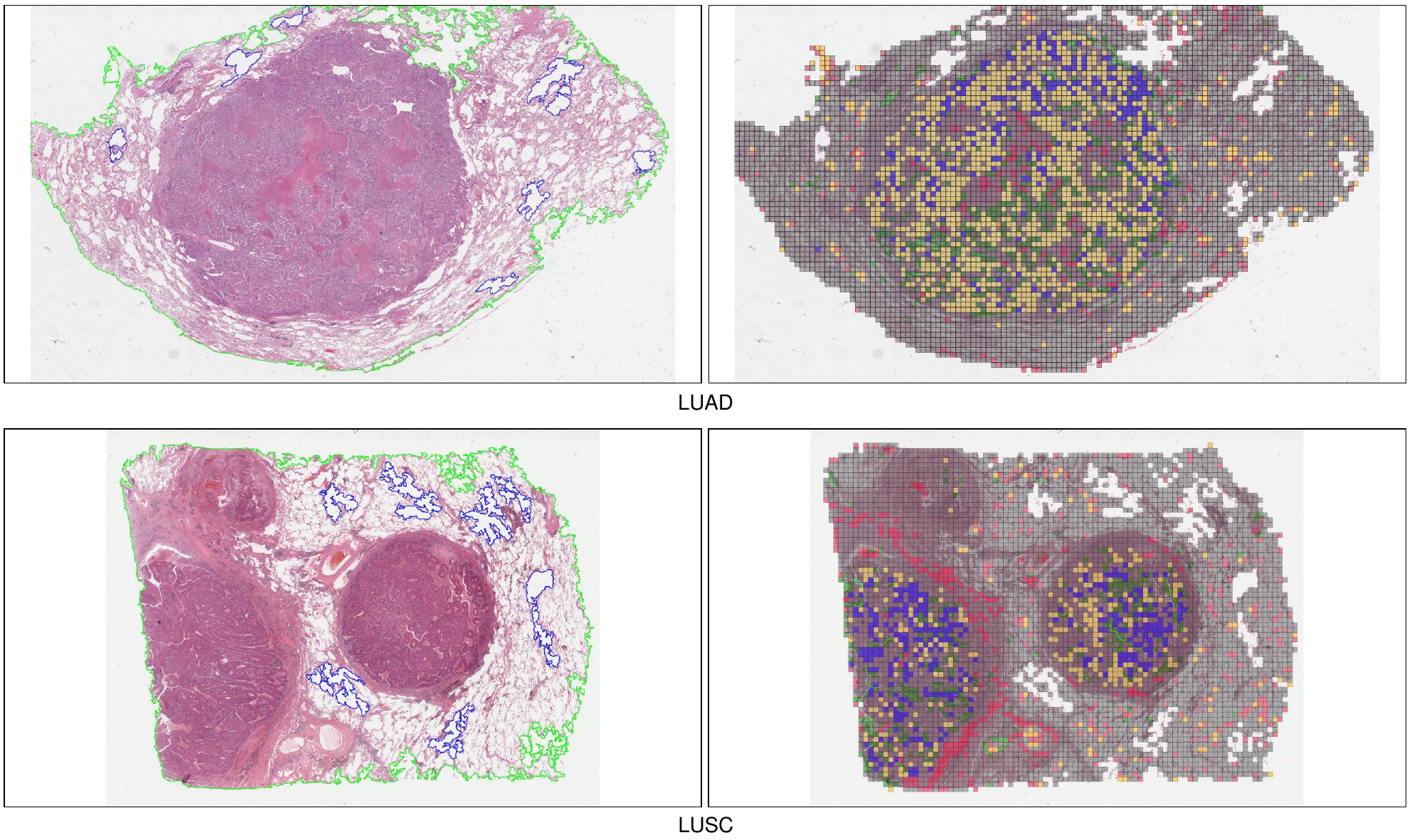}
\caption{Prototype visualization on Whole Slide Images (NSCLC).
}
\label{fig:nsclc}
\end{figure*}

\begin{figure*}[t]
\centering
\includegraphics[width=0.99\textwidth]{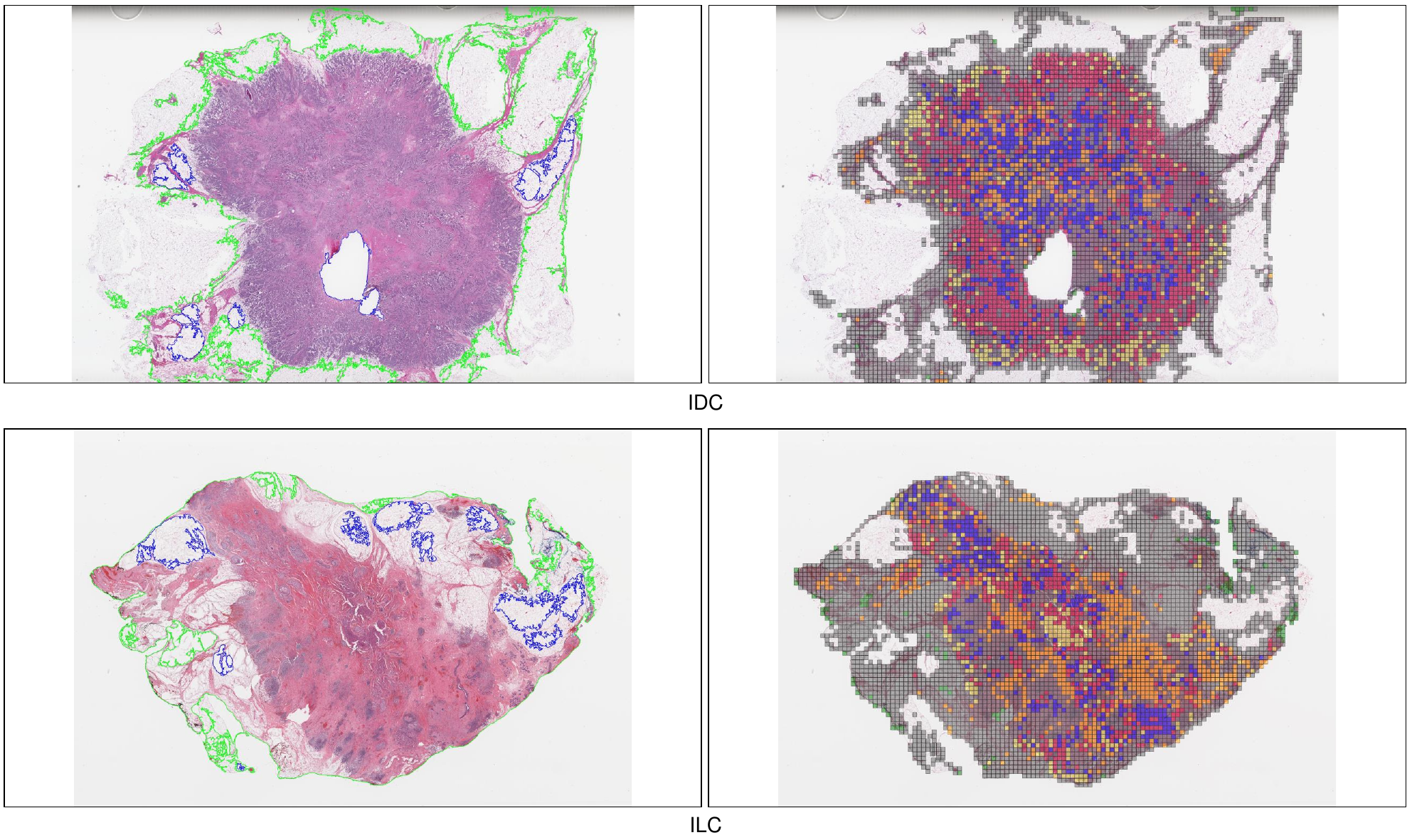}
\caption{Prototype visualization on Whole Slide Images (BRCA).
}
\label{fig:brca}
\end{figure*}

\begin{figure*}[t]
\centering
\includegraphics[width=0.99\textwidth]{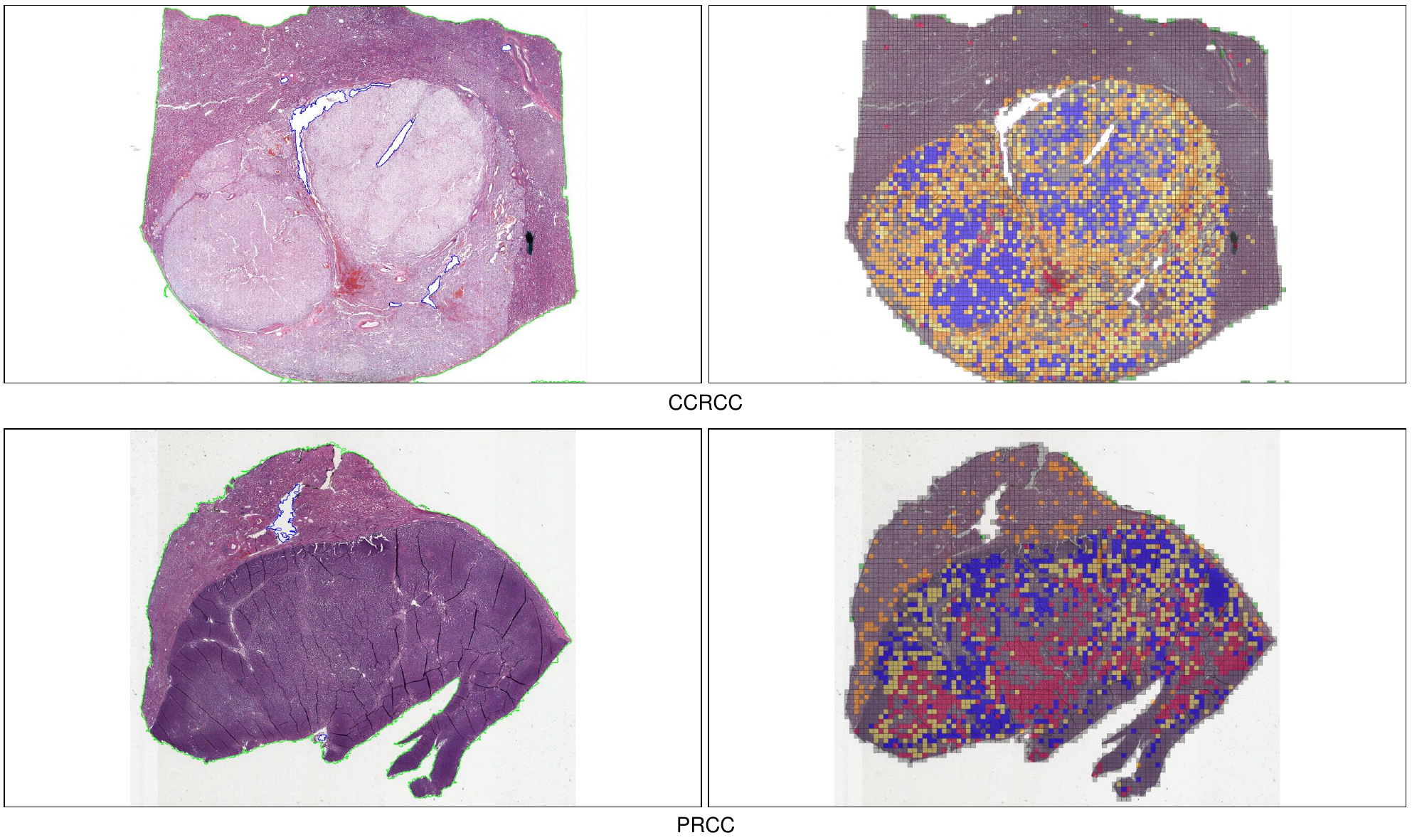}
\caption{Prototype visualization on Whole Slide Images (RCC).
}
\label{fig:rcc}
\end{figure*}

\begin{figure*}[t]
\centering
\includegraphics[width=0.99\textwidth]{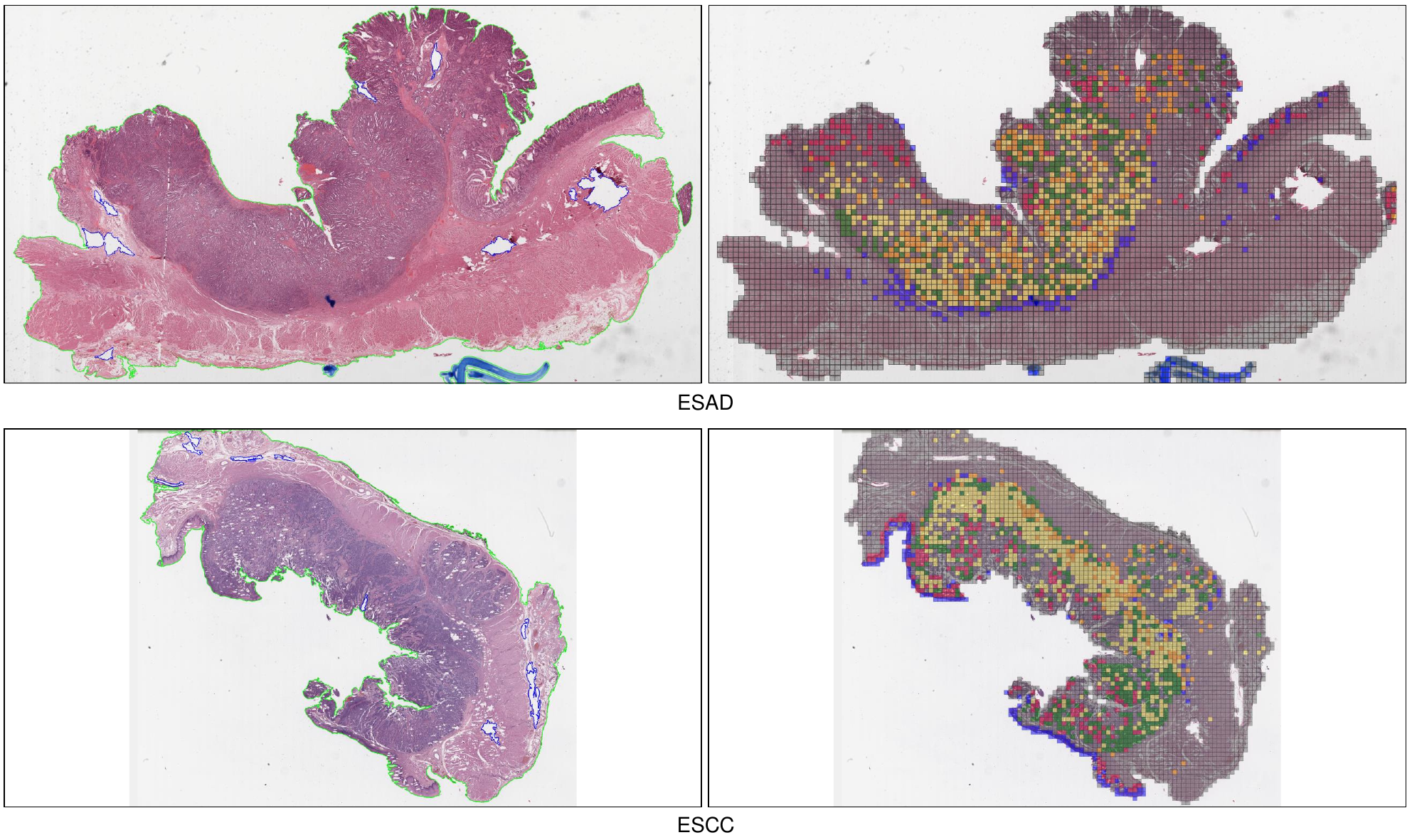}
\caption{Prototype visualization on Whole Slide Images (ESCA).
}
\label{fig:esca}
\end{figure*}

\subsection{G. More Analysis on Class Feature Enhancement}

Since in QPMIL-VL, class features guide the learning of WSI bag-level features, we further validate the effectiveness of the CFE module by visualizing the distribution of the bag-level features learned by the model before and after using the CFE module in four different incremental learning stages. The results are shown in Fig.~\ref{fig:bag_features}.

From these results, we could find that our CFE makes i) intra-class bag-level features more compact (dashed box) and ii) inter-class ones more distinguishable (solid box). This finding suggests that the class features used for supervising bag-level feature learning are improved.

\begin{figure*}[t]
\centering
\includegraphics[width=1\linewidth]{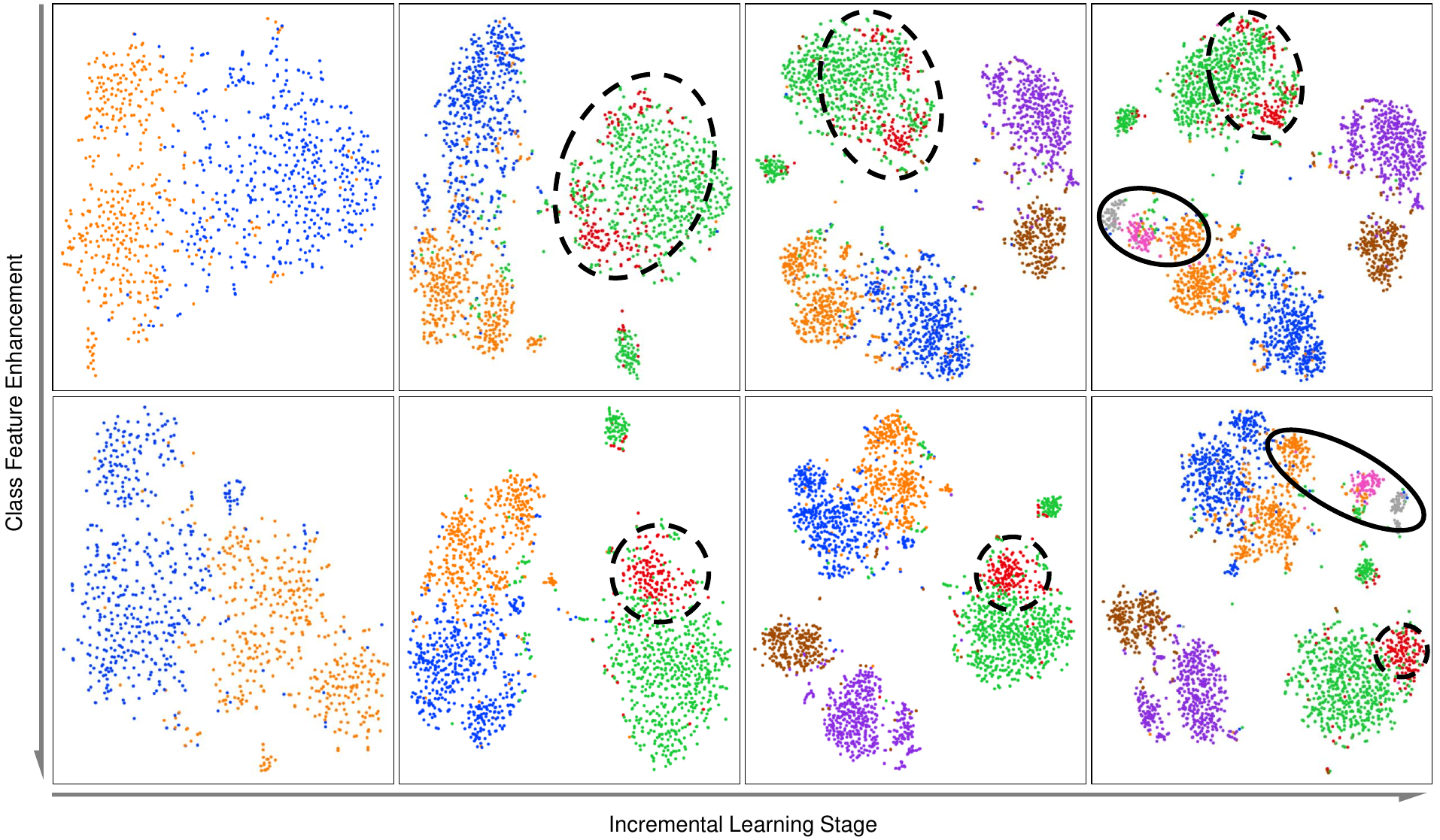}
\caption{Bag-level features visualization with or without our Class Feature Enhancement (CFE) in four incremental learning stages.
}
\label{fig:bag_features}
\end{figure*}

}

\end{document}